%% file: main.tex
\documentclass[11pt, a4paper, logo, one column]{craftjarvis}

\pdftrailerid{redacted}

\makeatletter
\renewcommand\bibentry[1]{\nocite{#1}{\frenchspacing\@nameuse{BR@r@#1\@extra@b@citeb}}}
\makeatother

\input{config}

\usepackage[authoryear, sort&compress, round]{natbib} 

\title{\cj: Open-world Multi-task Agents with Memory-Augmented Multimodal Language Models}

\reportnumber{} 

\renewcommand{\today}{} 

\author[1]{Zihao~Wang}
\author[1]{Shaofei~Cai}
\author[2]{Anji~Liu}
\author[3]{Yonggang~Jin}
\author[3]{Jinbing~Hou}
\author[1]{Bowei~Zhang}
\author[1]{Haowei~Lin}
\author[3]{Zhaofeng~He}
\author[4]{Zilong~Zheng}
\author[1]{Yaodong~Yang}
\author[4]{Xiaojian~Ma}
\author[1]{Yitao~Liang}

\affil[1]{PKU}
\affil[2]{UCLA}
\affil[3]{BUPT}
\affil[4]{BIGAI}
\affil[ \hspace{-0.73ex}]{All authors are affiliated with Team CraftJarvis\\} 

\begin{abstract}
    Achieving human-like planning and control with multimodal observations in an open world is a key milestone for more functional generalist agents. Existing approaches can handle certain long-horizon tasks in an open world. However, they still struggle when the number of open-world tasks could potentially be infinite and lack the capability to progressively enhance task completion as game time progresses. We introduce \cj, an open-world agent that can perceive multimodal input (visual observations and human instructions), generate sophisticated plans, and perform embodied control, all within the popular yet challenging open-world Minecraft universe. Specifically, we develop \cj on top of pre-trained multimodal language models, which map visual observations and textual instructions to plans. The plans will be ultimately dispatched to the goal-conditioned controllers. We outfit \cj with a multimodal memory, which facilitates planning using both pre-trained knowledge and its actual game survival experiences. 
    \cj is the existing most general agent in Minecraft, capable of completing over 200 different tasks using control and observation space similar to humans. These tasks range from short-horizon tasks, e.g., "chopping trees" to long-horizon tasks, e.g., "obtaining a diamond pickaxe". \cj performs exceptionally well in short-horizon tasks, achieving nearly perfect performance. In the classic long-term task of \texttt{ObtainDiamondPickaxe}, \cj surpasses the reliability of current state-of-the-art agents by 5 times and can successfully complete longer-horizon and more challenging tasks.
  The project page is available at \url{craftjarvis.org/JARVIS-1}.
\end{abstract}

\begin{document}

\correspondingauthor{Xiaojian~Ma,~Yitao~Liang\\
Zihao Wang<zhwang@stu.pku.edu.cn>, Shaofei Cai<caishaofei@stu.pku.edu.cn>, Anji Liu<liuanji@cs.ucla.edu>, Xiaojian~Ma<xiaojian.ma@ucla.edu>, Yitao Liang<yitaol@pku.edu.cn>}

\maketitle

\section{Introduction}

\begin{figure*}[t]
    \centering
    \includegraphics[width=\linewidth]{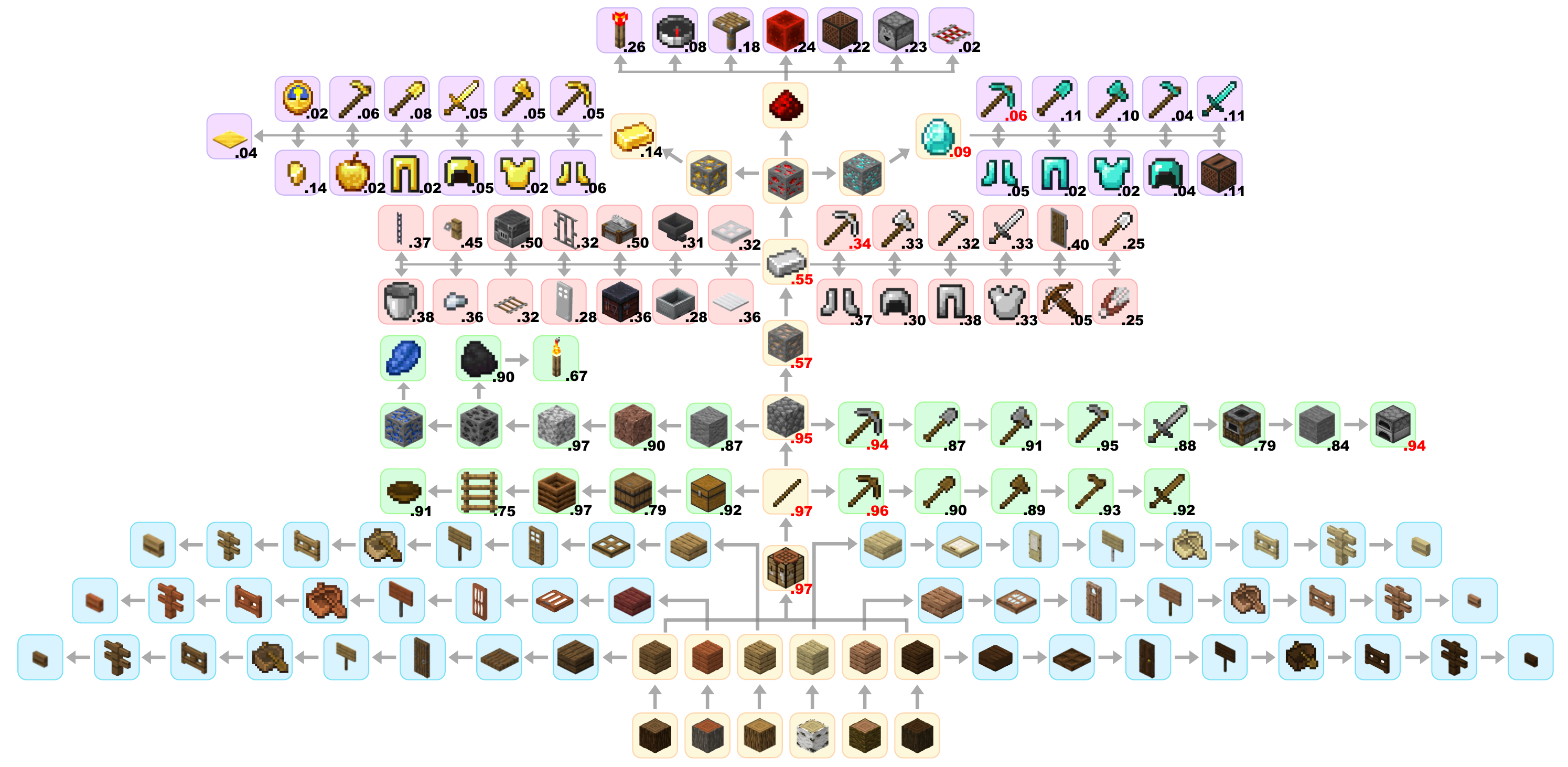}
    \caption{
        \textbf{How does \cj unlock the technology tree of the Minecraft universe.} 
        \cj can consistently obtain high-level items on the main tech-tree of the overworld in Minecraft, such as diamond, redstone, and golden items, which require collecting over 10 different intermediate items. \cj not only outperforms the previous state-of-the-art VPT~\citep{vpt} (6\% vs. 2.5\% reliability) on diamond pickaxe, but also can craft almost all diamond items in the overworld including diamond chestplate.}
    \label{fig:tech_tree}
\end{figure*}

Creating sophisticated agents that can accomplish myriad of tasks in complex domains remains a pivotal milestone towards generally capable artificial intelligence~\citep{gato,gpt3,flamingo,rt1,jxma_vl1}. Recent advancements have shown a trend towards employing a hierarchical goal execution architecture~\citep{deps,innermonologue,huang2022language}, and leveraging large language models (LLMs) as the high-level planner to generate action plans that will be ultimately executed by low-level instruction-following controllers. Albeit the fruitful progress they have yielded in many robotics~\citep{innermonologue} and even open-world environments like Minecraft~\citep{minedojo,minerl}, today's agents built with these approaches are still struggling with three major issues: 1) perceive the world from multimodal sensory observations, such as images, videos in addition to natural language instructions and feedback for planning; This is mostly due to the inability of LLM-based planners on processing multimodal data~\citep{huang2022language,react}; 2) perform consistent and accurate long-term planning. This requires multi-round, knowledge, and reasoning-intensive dialogues, which remain great challenges to LLMs~\citep{innermonologue}; 3) learn and evolve in a life-long fashion. This calls out the need for agents to propose their own tasks and self-improve. Addressing these issues will unleash the full planning potential of LLM-based agents, and expedite the development of more generalist agents.

In this work, we introduce \cj, a brand new agent that can robustly produce plans for long-horizon tasks from multimodal user and environment inputs, and translate them into motor control in Minecraft, a popular yet challenging open-world testbed for generalist agents. To be specific, we chain a multimodal foundation model MineCLIP\!~\citep{minedojo} and an LLM\!~\citep{gpt3} together, the resulting multimodal language model (MLM) allows our agent to better understand the task, situations, and environmental feedback. To further enhance the correctness and consistency of planning, especially on long-horizon tasks, we propose to augment the agent with a \textit{multimodal memory}, which stores both the scenarios and actual plans of the successful planning experiences in the past. By retrieving the relevant memory entries, the planning skill of our MLM-based agent can be strengthened from the agent's own interactions with the environment in an in-context manner. Finally, \cj is able to evolve throughout the gameplay by proposing tasks on its own (\ie self-instruct) as a means of exploration and saving the obtained experiences in the multimodal memory, therefore facilitating better reasoning and planning. 
This self-improving ability sparks its potential for a higher level of autonomy.  

Our main evaluations are conducted in Minecraft, with more than 200 tasks selected from the Minecraft Universe Benchmark~\citep{mcu}, with no demonstration provided. 
The tasks cover a broad spectrum from the early game (\eg \texttt{ObtainCraftingTable}) to intermediate and even challenging long-horizon tasks (\eg \texttt{ObtainDiamondPickaxe}). A glimpse of what \cj is able to achieve can be found in Figure~\ref{fig:tech_tree}. 
\cj exhibits strong performances on these tasks, representing an up to 5$\times$ increase to the previous records. 
Our ablative analysis then offers a detailed account of how \cj approaches this significant progress and becomes the first agent that can robustly obtain the diamond pickaxe with up to 12.5\% success rate. What is even more surprising is that, without the need for additional training, \cj demonstrates a continuous increase in performance as game time increases in long-horizon tasks.
Moreover, \cj has demonstrated its potential of \textit{self-improve} in an exploratory life-long learning experiment, where it needs to propose tasks to progressively explore the world, collect experiences, and sharpen its planning skill using these experiences stored in the multimodal memory. 

\noindent
In summary, \cj pilots the effort towards a human-like multi-task and autonomous agent in an open-world, embodied environment like Minecraft. We would like to share the key takeaways of what we have learned during its development as follows:

\begin{itemize}
 \item \textbf{From LLMs to MLMs.}~~The capability of perceiving multimodal sensory input is critical to planning in a dynamic and open-world world. \cj enables this by chaining a multimodal foundation model together with an LLM. Compared to LLM ``blindly'' produces plans, MLM is able to natively understand the current situation and plan accordingly. Further, rich environmental feedback can be obtained through multimodal perception, therefore helping the \textit{self-check} and \textit{self-explain} of the planner spot and fix possible bugs in the plans, enabling stronger interactive planning. 
 \item \textbf{Multimodal memory.}~~Early research has suggested the crucial role that memory mechanisms can serve in the functioning of generalist agents. By outfitting \cj with a multimodal memory, we effectively allow it to plan with both pretrained knowledge and its actual experiences in the world, therefore bringing significant improvement to planning correctness and consistency. Compared to canonical RL or planning agents with exploration, no additional model update is needed as the MLM in \cj makes it possible to leverage these experiences in an in-context manner. 
 \item \textbf{Self-instruct and self-improve.}~~A sign of generalist agents is the capacity to proactively acquire new experiences and continuously improve themselves. We have demonstrated how \cj effectively traverses the environment by executing tasks autonomously generated through its \textit{self-instruct} mechanism. With multimodal memory teaming up with experiences from the explorations, we have observed consistent improvement, especially in accomplishing more complicated tasks. Ultimately, this aspect of autonomous learning in \cj signifies an evolutionary step towards generalist agents that can learn, adapt, and improve over time with minimal external intervention.
\end{itemize}

\section{Challenges for Open-world Agents}

Compared to canonical scenarios with relatively small scale, simple dynamics, and limited tasks, open-world environments impose substantial challenges to building agents that can accomplish a diverse set of tasks~\citep{minedojo,minerl-rel-1,minerl-rel-2,minerl-rel-3,shaofei,deps,groot}. In this section, we will review three major challenges we've identified during the development of \cj.

\subsection{Challenge I: Situation-Aware Planning}\label{sec:challenge_sit_planning}

\begin{figure*}[t]
    \centering
    \includegraphics[scale = 0.55]{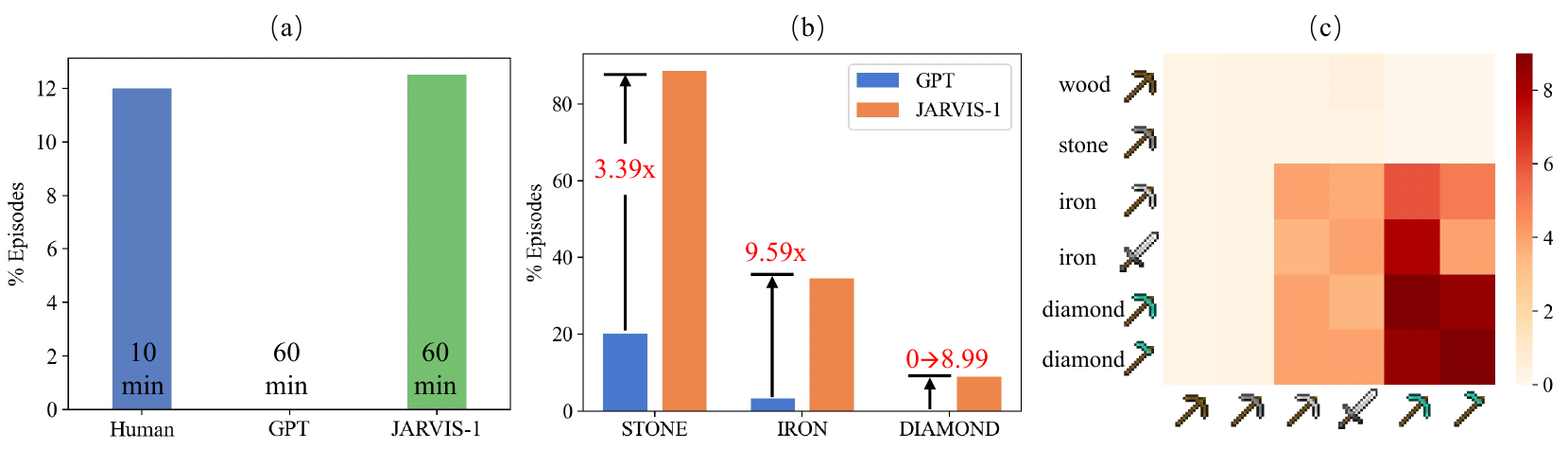}
    \caption{\textbf{Challenges in open-world environments and how does \cj tackle them}. \textbf{(Left)} With situation-aware planning, \cj substantially improves the success rate on the challenging \texttt{ObtainDiamond} task, compared to the baseline (GPT) without it. Note: Due to resource constraints, we can only provide human results of 10-min gameplay; \textbf{(Middle)} As task complexity increases (STONE$\rightarrow$IRON$\rightarrow$DIAMOND), \cj exhibits more significant advantages thanks to interactive planning; \textbf{(Right)} Success rate gain (indicated by the color depth) on selected tasks (x-axis) given in-context experiences on other tasks (y-axis) retrieved from the multimodal memory. With life-long learning and memory, \cj can utilize prior experiences on relevant tasks for better planning. 
    }
  \label{fig:challenge}
\end{figure*}

In an open world, there could be various possible paths towards an open-world goal. However, not all of them are plausible or equally efficient given a certain \textit{situation} (location, inventory status, \etc).
For example, building a bed \includegraphics[scale=0.04,valign=c]{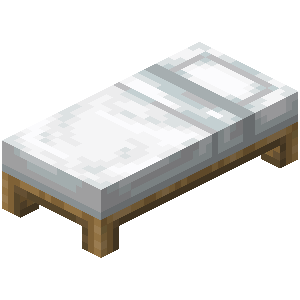} can be done through collecting wool from sheeps \includegraphics[scale=0.16,valign=c]{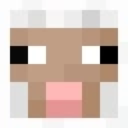}, haunting spiders for strings \includegraphics[scale=0.16,valign=c]{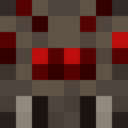}, or trading with villagers \includegraphics[scale=0.015,valign=c]{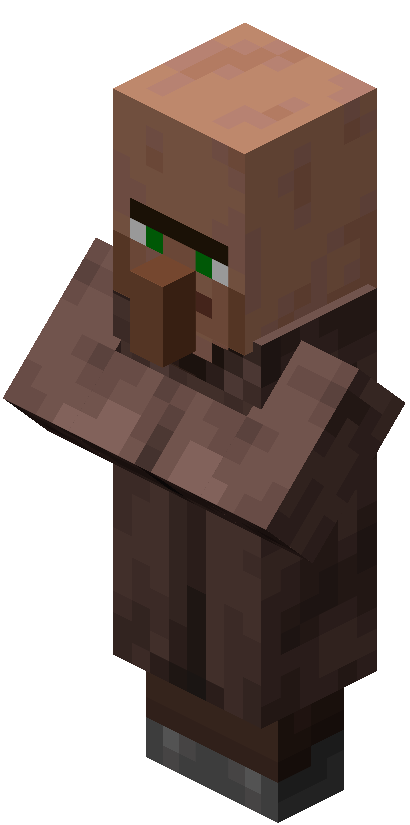}. Depending on the current location and its proximity to these subjects, some options can be more viable and more efficient than others. Further, the agent's own situation can also change throughout the episode, \eg day and night shifts, weather conditions (bringing different types of danger), and tool usage (it can be broken). To this end, the plan needs to be constantly updated based on the current situation.
Figure~\ref{fig:challenge} (left) shows that when attempting the "ObtainDiamondPickaxe" task with a GPT-based planner that produces plans only at the beginning without looking at the current situation, the agent failed to complete the task as opposed to human players and \cj, which perform situation-aware planning from time to time. We've observed that many failures coming from this were attributed to the agent's inability to adapt to the changing situations including entering a new biome, the tool being used becoming broken, \etc. 

\subsection{Challenge II: Task Complexity}\label{sec:challenge_task_complexity}
The second challenge comes from the higher task complexity in open-world environments.
Due to the richness of terrains, objects, and action space, tasks in open-world domains usually require substantially long planning horizons as well as good accuracy and precision. For example, the task \texttt{ObtainEnchantingTable} \includegraphics[scale=0.04,valign=c]{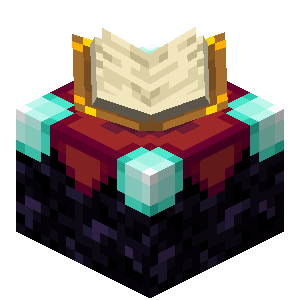} includes more than 20 different sub-goals and therefore demands significantly longer reasoning steps. Meanwhile, many of these sub-goals have to be achieved precisely with the exact object name, quantities, and preconditions, e.g., \texttt{mine 3 obsidian with diamond pickaxe}, \texttt{craft 1 diamond pickaxe from 3 diamonds and 2 sticks}; otherwise, the subsequent sub-goals won't be executed due to unfulfilled preconditions. To tackle this, we may refer to some approaches in LLM reasoning, \eg self-debugging~\citep{selfdebug} and turning the planning into an interactive fashion.
In \autoref{fig:challenge} (Middle), we've shown that as the complexity of the task increases, our \cj, which uses interactive planning~\citep{deps} to mitigate the aforementioned issues (details can be found in \autoref{sec:interactive_planning}), elicits more significant advantages over the baseline (GPT) planner. 

\subsection{Challenge III: Life-long Learning}\label{sec:challenge3}

Finally, being open world often implies offering an infinite number of tasks. Clearly, it is difficult for an agent to master all tasks or generalize to arbitrary tasks without additional learning. To this end, agents in an open world should be able to learn novel tasks while completing existing tasks, \ie life-long learning. Furthermore, as many open-world agents employ large models~\citep {deps,plan4mc,voyager,gitm}, canonical gradient-based learning could be extremely inefficient given the number of new tasks and experiences to learn. Our MLM-based \cj tackles this by adopting a memory to save all the experiences on past tasks. By retrieving memory entries relevant to the newly-coming task and putting them into the context as a reference, \cj is able to accumulate more experiences as the game continues and strengthen its own planning skills without gradient update. As illustrated in \autoref{fig:challenge} (Right), for instance, both \texttt{ObtainDiamondPickaxe}\includegraphics[scale=0.08,valign=c]{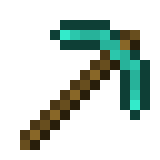} and \texttt{ObtainDiamondAxe}\includegraphics[scale=0.08,valign=c]{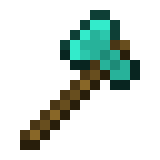} require gathering almost identical materials. Therefore, they can help each other by using the experiences from the other task. Compared to completing these challenging tasks without any prior experiences, memory-based in-context life-long learning in \cj can bring significant advantages.

\section{Multi-task Agent with Memory-Augmented MLM}

\begin{figure*}[t!]
    \centering
    \includegraphics[scale = 0.45]{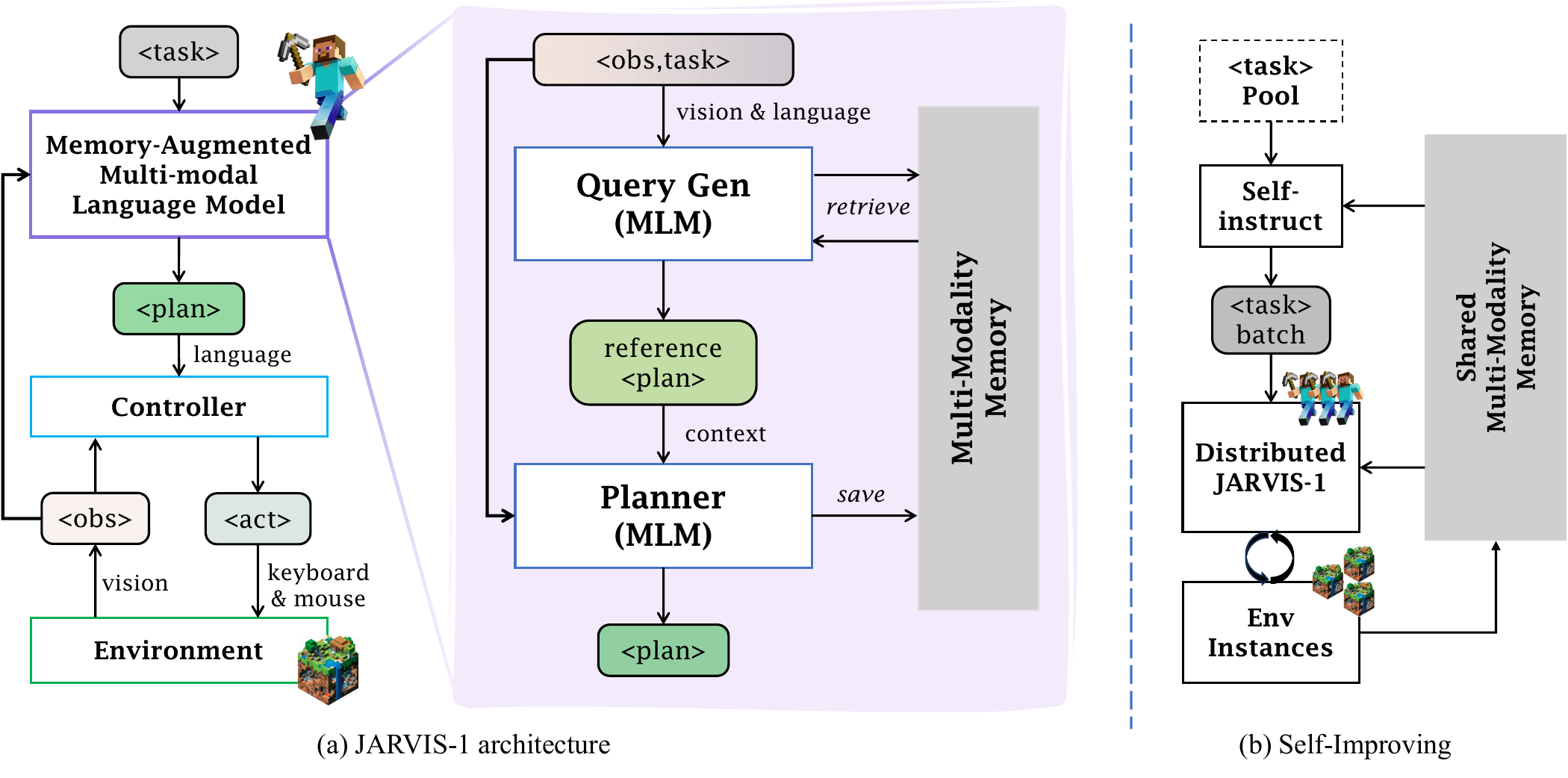}
    \caption{
        \textbf{Architecture of \cj and its self-improving mechanism}. 
        (a) \cj comprises a memory-augmented multimodal language model (MLM) that produces plans and a low-level action controller. \cj also utilizes a multimodal memory to store and obtain experiences as references for planning. (b) \cj can strengthen its own planning skills through exploration with its own proposed tasks (\textit{self-instruct}) and a growing memory that helps with better planning on tasks that has been (partially) visited before.
        }
    \label{fig:pipeline}
\end{figure*}

This section details the architecture of the proposed \cj agent. We begin with an overview of the modular agent design in \autoref{sec:overview}. Next, we elaborate on how to implement an interactive planning scheme with a multimodal language model, which helps with more accurate plans, especially on complex and long-horizon tasks in \autoref{sec:interactive_planning}. Finally, we show how to augment this planning framework with a multimodal memory to allow \cj to strengthen its planning skill throughout the episode by in-context life-long learning in \autoref{sec:planning_with_memory} and \autoref{sec:learning}.

\subsection{Overview}\label{sec:overview}

We aim to develop an agent capable of solving long-horizon instruction-following tasks using image observations and human-aligned actions. To accomplish this, we propose a multi-modal agent including an interactive \textbf{planner}, a goal-conditioned \textbf{controller}, and a \textbf{multimodal memory} of multimodal experiences. Upon receiving a task and the current observation, \cj first utilizes the MLM to generate a multimodal query (\textbf{query gen}) that retrieves relevant planning experiences from the memory. These experiences will then be used along with the planning instruction to prompt the MLM-based planner. Leveraging its own pretrained knowledge as well as the retrieved reference plans, the planner will ultimately produce a series of $K$ short-horizon goals $g_1,\ldots,g_K$ to be executed by the controller. Once the plan is successfully executed, it will be stored in the memory along with the task and the agent situation when it was planned. We also empower \cj with life-long learning by combining \textit{self-instruct}, where \cj will propose some tasks for itself to complete as a means of exploration; and \textit{self-improve}, where multiple \cj agents will be running in parallel to gather experiences, therefore helping with better planning later. We provide an illustration in \autoref{fig:pipeline}.

\subsection{Interactive Planning with MLM}\label{sec:interactive_planning}

\begin{figure*}[t!]
    \centering
    \includegraphics[scale = 0.45]{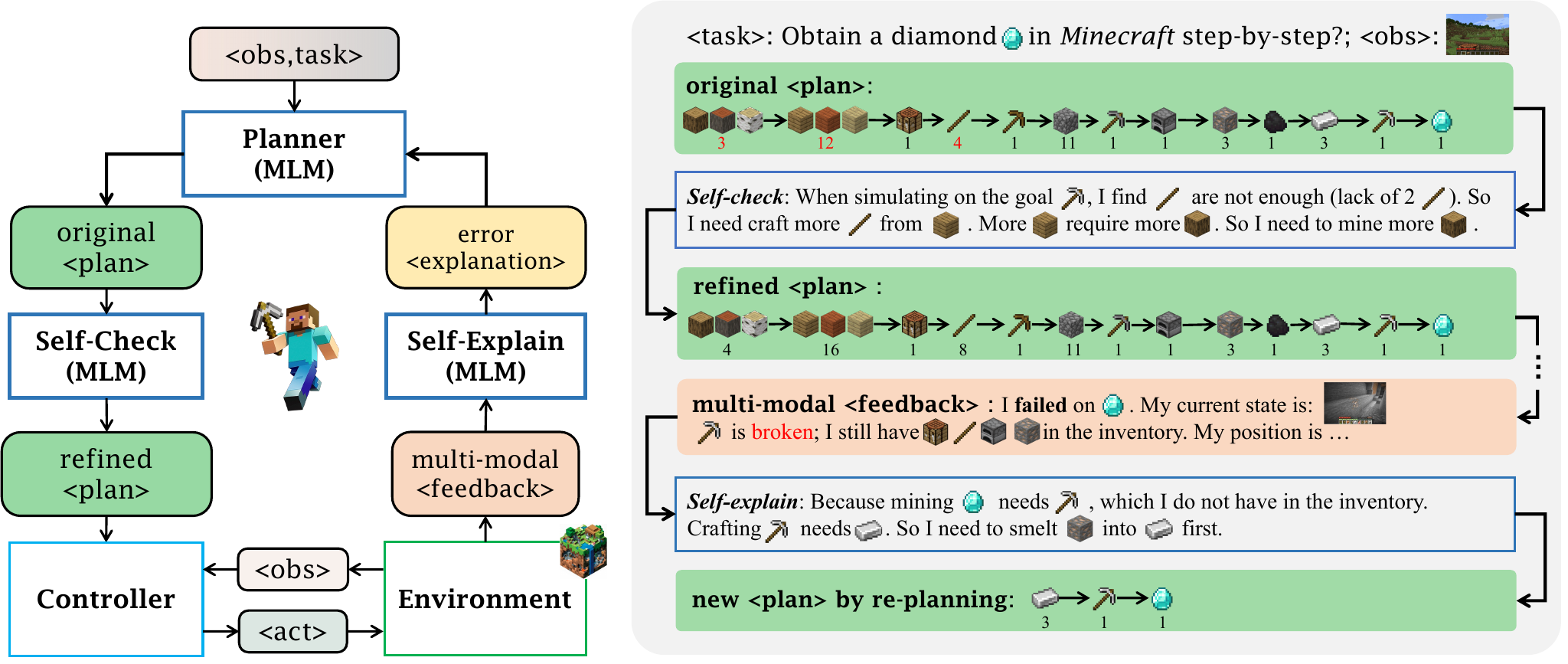}
    \caption{
        \textbf{Interactive planning in \cj}. After receiving the current task instruction and observation, \cj will produce an initial plan, which will go through \textit{self-check} to get possible bugs (marked in {\color{red}{red}}) fixed. Further, in case any error (also marked in {\color{red}{red}}) occurs during the execution of the refined plan, \cj will try to reason about the next move from the environmental feedback via \textit{self-explain}. Interleaving self-check and self-explain significantly boosts the correctness and robustness of \cj planning.
        }
    \label{fig:planner}
\end{figure*}

As we have mentioned in \autoref{sec:challenge_sit_planning} and \autoref{sec:challenge_task_complexity}, the primary challenges for planning in Minecraft come from the requirement of being able to plan for long-horizon tasks under dynamic observations. Confirmed by many prior arts~\citep{deps,voyager,plan4mc}, this makes it exceptionally hard to utilize canonical symbolic planners, which can be much less flexible. To this end, we take a multimodal language model (MLM) as zero-shot \textbf{planner} and combine it with an interactive planning framework to tackle these challenges.

\textbf{Situation-aware planning with MLM.}~~To achieve situation-aware planning, the planner must take the current observation into account, in addition to the task instruction~\citep{huang2022language,react}. Specifically, we begin with translating the multimodal observation into text descriptions. As opposed to letting the MLM caption the scene directly, we first extract keywords of Minecraft items (e.g., "\texttt{acacia tree}", "\texttt{sheep}") from Minecraft wiki and utilizing GPT~\citep{gpt3} to generate sentences that describe these observations. For example, a generated sentence could be "I can see sheep in the acacia plains". Then the MLM will retrieve the condition sentence according to current visual observation during planning. Additional situation details including biome and inventory status are also converted into text using templates. Finally, we prompt the MLM again (the language part only) into a plan given the task instruction and all the aforementioned textual situation descriptions. Compared to end-to-end alternatives~\citep{brohan2023rt,leo}, we find our composable usage of MLM provides higher quality situation descriptions and ultimately, plans with much less hallucination.

\textbf{Planning with self-check.}~~Our first layer of shield to ensure the correctness of plans involves \textit{self-check}. Similar to self-debugging\!~\citep{selfdebug}, given an initial plan, we ask \cj to progressively simulate the plan execution, predict the resulting state after each step (primarily the state of inventory), and evaluate them. By verifying if these states satisfy the goal's precondition, \cj can proactively identify potential plan flaws. Compared to the canonical planner where the agent has to encounter the error first before making a remedy, this upfront plan verification could mitigate the need for the agent to recover (re-plan) from more challenging situations due to plan failure. For instance, if an agent starts digging underground without sufficient wood, it would typically have to return to the surface, which substantially lowers the chance of completing the task.

\textbf{Planning with environment feedback.}~~Next, our interactive planning framework ventures into allowing \cj to quickly recover from failure by leveraging environment feedback in a closed-loop fashion. The process is illustrated in \autoref{fig:planner}. During plan execution, we feed the feedback to the MLM of \cj in case there is any execution failure (possibly due to a flawed plan) and utilize its self-explain mechanism~\citep{reflexion} to explain the error and locate the bugs in the original plan (we term this as \textit{error explanation}). Finally, the MLM planner of \cj will produce an improved plan based on both the outside environment feedback and the inside retrospective. Compared to other agents that rely on human intervention or privileged environment information~\citep{innermonologue,gitm}, \cj has the ability to speculate about the reasons why current goals cannot be achieved, without the need for additional information or design.

\subsection{Planning with Multimodal Memory in the Loop}\label{sec:planning_with_memory}

To address the life-long learning challenge mentioned in \autoref{sec:challenge3}, we equip \cj with multimodal memory to allow learning from its own past experiences. We will detail the formulation of the retrieval-augmented planning, query generation, and memory layout below.

\textbf{Retrival-augmented planning.}
Retrieval-augmented generation (RAG)~\citep{lewis2020retrieval,mao2020generation} enhances the quality of responses generated by LLMs by incorporating external sources of knowledge to complement the model's internal representation.
We also utilize RAG to enhance \cj's long-term planning capability.
Compared to official RAG methods leveraging the external knowledge library, we take the collected multimodal memory as the knowledge library and retrieve the interactive experiences as the demonstration prompt to augment the planning results. 
The formulation is as follows:
\begin{align}
    p(y \mid x) \approx  \sum_{z\in \text{top-k}(p(\cdot \mid x))}p_\eta (z \mid x) p_\theta (y \mid x,z),
\end{align}
where $x$, $y$, and $z$ denote instruction, plans, and retrieved memory entries respectively, and $p_\eta$ and $p_\theta$ are denoted as retrieval and planning models.
Such retrieval-augmented planning method helps \cj ground the internal knowledge into the open-ended environments efficiently and leverage the historical interaction feedback to solve the hallucination within LLMs and produce more accurate plans.

\begin{figure*}[t!]
    \centering
    \includegraphics[scale = 0.5]{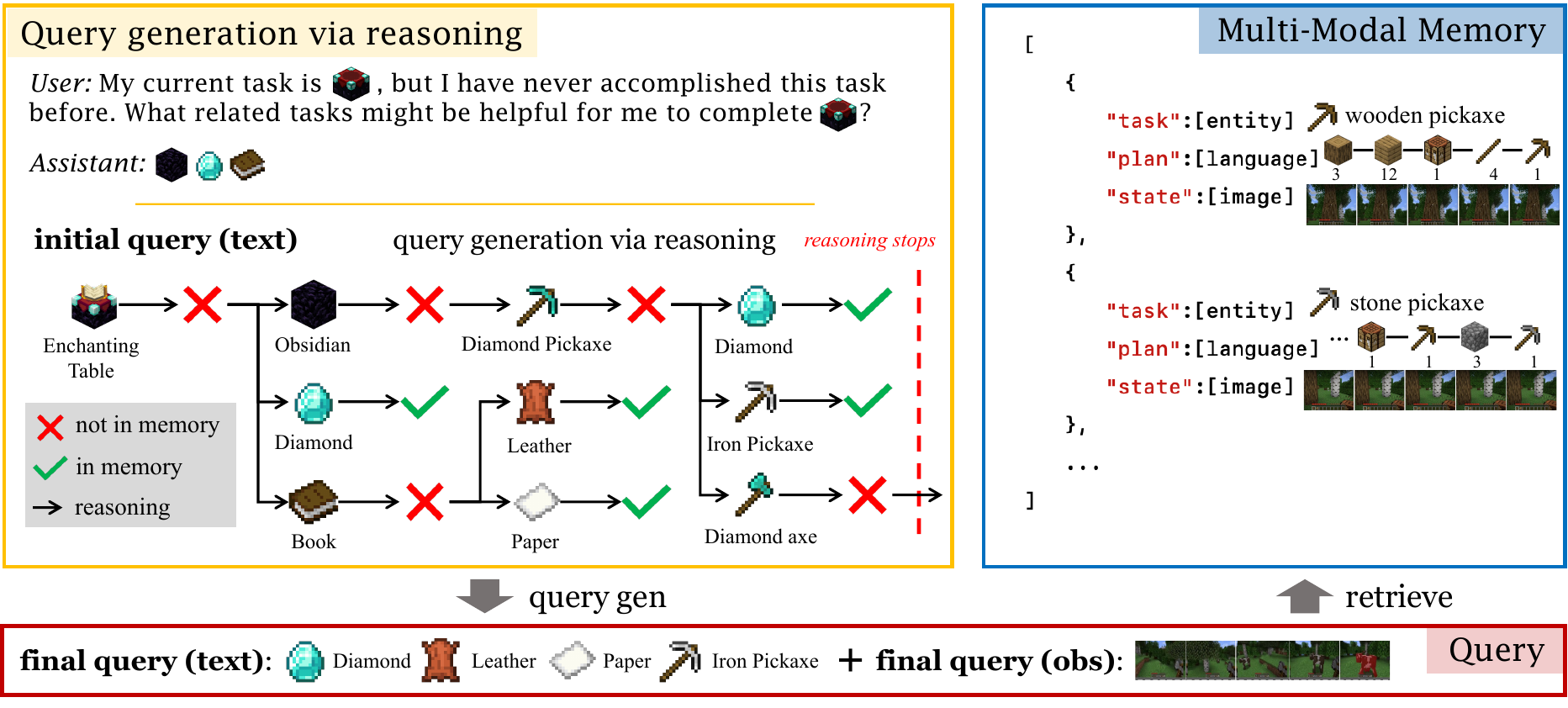}
    \caption{\textbf{Query generation in \cj}. Given the current observation and the task, \cj will first think backward and figure out the needed intermediate sub-goals. The reasoning will be bounded by a limited depth. The sub-goal that is present in the memory will join the current visual observation to form the final query. Entries that match the text query will be ranked by the perceiving distance of their states to the obs query and only the top entry of each sub-goal will be retrieved.
        }
    \label{fig:memory_retrieval}
\end{figure*}

\textbf{Multimodal memory.}~~We have demonstrated the layout of our multimodal memory on the right side of Figure~\ref{fig:memory_retrieval}. From a high level, it is a key-value memory where the keys are multimodal, comprising both the task and the observation (or situation) made when this memory entry was created. The values are the plans that were successfully executed. Note that, since the plans in an open-world environment like Minecraft are situated (see \autoref{sec:challenge_sit_planning}), there could be multiple entries that are with the same task but different observations and plans. As a result, \cj needs to produce multimodal queries based on the current task and situations to retrieve the relevant memory entries.

\textbf{Query generation via reasoning.}~~ 
When presented with an instruction as a task, we employ query generation via LLM reasoning to decompose the instruction into sub-tasks or related tasks, which will then be used as textual queries to retrieve relevant planning experiences as references for solving the current task. For instance, consider the instruction "craft 1 enchanting table with empty inventory" as shown in Figure~\ref{fig:memory_retrieval}. \cj queries the MLMs to identify the tasks that are required for achieving the main task in a backward search fashion, e.g., ``obtain book \includegraphics[scale=0.03,valign=c]{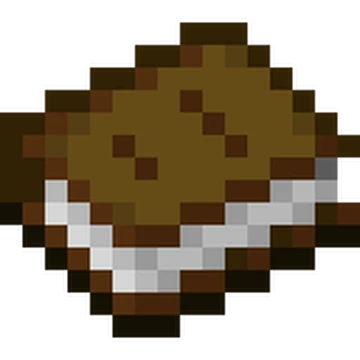}/diamond \includegraphics[scale=0.04,valign=c]{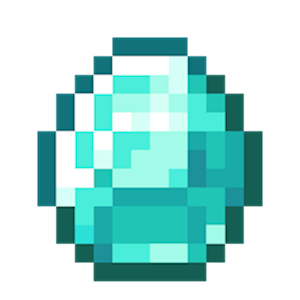}/obsidian \includegraphics[scale=0.03,valign=c]{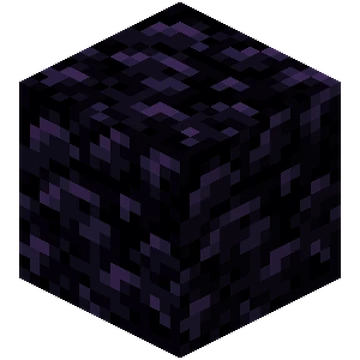} with empty inventory''. The search depth is bounded for efficiency.
Further, instead of relying solely on retrieval based on the text query~\citep{voyager,gitm}, we also propose to append the agent's current visual observation to the textual query, resulting in a \textit{multimodal query} to take the situation into account during memory retrieval.

\textbf{Multimodal retrieval.}
After obtaining the textual and visual query, we compute the alignment between the query and each trajectory in multimodal memory.
We first use the text encoder of the CLIP model to compute the embedding of the query and task key of each entry in memory. We select the memory entries with similarity higher than the confidence threshold as the candidate entries. 
Then we will compute the visual state embedding of query and states in candidate entires. Then we sort the candidate entries with the visual embedding similarities, which can be formed as: 
\begin{align}
    p_\eta(z\mid x) \propto & \text{CLIP}_{v}(s_z)^\top \text{CLIP}_{v}(s_x),
\end{align}
where $s_z$ and $s_x$ are the visual key of memory entries and visual query, respectively. Finally, we retrieve the plan of top-k candidate entries as reference prompt $z$.

\subsection{Self-improving Agents}\label{sec:learning}

\textbf{Learning in Minecraft with memory.}~~The remaining issue now is where the aforementioned multimodal memory comes from. Inspired by the life-long learning scheme in many close-world and open-world reinforcement learning problems~\citep{lifelong1,lifelong2,voyager}, we propose the following learning approach for augmenting the memory in \cj: 1) First, we generate a set of tasks, which form some curricula for the agents to complete as means of exploration of the world. During this process, \cj produces plans, interacts with the environment, embraces the errors, and stores all these experiences in the memory; 2) After this learning stage, we evaluate \cj on various tasks. Therefore, \cj is able to produce better plans with the memory teaming up with the planning experiences. In our experiments, we use this as the default setting for all tasks.

\textbf{Exploration using self-instruct.}~~The key issue to the success of learning with memory is how to effectively acquire useful experiences given a limited amount of time. We propose to use self-instruct~\citep{selfinstruct} to generate the dynamic curriculum and guide \cj to learn from the interactions with environments. In each round, we prompt the MLM to consider how capable \cj is at this point and subsequently select tasks from a task pool to explore. We find that the curriculum almost follows the technical tree-growing direction. To accelerate the learning process, we augment the linear self-instruct to distributed learning in distributed environments with shared memory, \ie speculative execution~\citep{leviathan2023fast}. Specifically, we generate multiple executable tasks as candidate task batches and provide them to agents with the same memory for verification and execution in various different environments. Meanwhile, experiences are collected into a shared centralized memory. When all exploration tasks have been accomplished, we move to the next round, until the memory reaches a certain capacity.

\textbf{Life-long learning.}~~We've also observed that the aforementioned learning (where the memory is being filled) can be extended throughout the whole gameplay, where the agent gradually acquires more and more skills. As the gameplay continues, more and more experiences are pouring in, therefore \cj can find better references for challenging tasks like \texttt{ObtainDiamondPickaxe}, resulting in an improved success rate on these tasks. Further, there is no gradient update in this thanks to the memory-augmented MLM, \ie we can do in-context life-long learning. In Section~\ref{sec:long_horizon_experiments}, we offer exploratory experiments to show the potential of such capability of \cj.

\section{Experiments}

In the experiments, our goal is to 1) evaluate the general performances of \cj on the challenging Minecraft tasks, especially on its advantages over baselines that do not (fully) address the aforementioned issues in open-world agents; 2) understand the factors that contributes to the general results; 3) explore the potential of \cj in terms of life-long learning and its benefits to long-horizon tasks. To this end, we will first briefly introduce the evaluation settings, then cover the main comparative results and ablation studies, and conclude with an exploratory trial on long-horizon tasks.

\subsection{Experimental Setups}\label{sec:task_setting}

We evaluate \cj in Minecraft, with tasks selected from the recently introduced Minecraft Universe Benchmark~\citep{mcu}. For the reader's convenience, we provide details on the basic setups below.

\textbf{Environment setting.}~~To ensure realistic gameplay, the agent needs to utilize observation and action spaces that are similar to those used by humans.
Instead of manually designing a custom interface for models to interact with the environment, as done in previous methods such as MineDojo\citep{minedojo}, GITM\citep{gitm}, and Voyager\citep{voyager}, we opt for using the native human interface provided by Minecraft. This applies to both the observation and action space.
The model operates at a speed of 20 frames per second and is required to use a mouse and keyboard interface when interacting with human GUIs.
For more information on the detailed descriptions of the observation and action spaces, please refer to the Appendix.

\textbf{Task setting.}~~
In Minecraft, players have access to thousands of items, each with specific acquisition requirements or recipes. For example, stone-type items can only be obtained using a pickaxe, and two planks can be crafted into four sticks (these requirements are available on the Minecraft Wiki$^1$). In survival mode, players must obtain each type of item from the environment or craft/smelt the object item from materials. 
We choose over 200 tasks from the Minecraft Universe Benchmark~\citep{mcu} for evaluation. These tasks are related to items that can be obtained in the Minecraft overworld.
For the convenience of statistics, we have classified them into 11 groups according to recommended categories in Minecraft$^2$ (see Table\ref{tab:mc_task_set}).
Due to the varying complexity of these tasks, we adopt different maximum gameplay durations (Max. Steps) for each task. The limit is determined by the average time the human players need to accomplish the corresponding task.
Other details about each task, such as language instruction, maximum steps, evaluation times, biome, and initial inventory when the agent is born into the world can be found in Appendix Table~\ref{tab:wood_results}-\ref{tab:food_results}.

\fancyfoot[L]{
\footnotesize{$^1$https://minecraft.fandom.com/wiki/Minecraft\_Wiki}
\footnotesize{$^2$https://minecraft.fandom.com/wiki/Tutorials/Organization\#Categories}
}

\input{tables/mc_task_set}

\textbf{Evaluation metrics.}~~
By default, the agent always starts in survival mode, with an empty inventory. A task is considered a success when the target object is obtained within a specified time. Due to the open-world nature of Minecraft, the world and initial position that the agent is spawned at could vary a lot. Therefore, we conducted at least 30 tests for each task using different seeds and reported the average success rate to ensure a thorough assessment. Further, since we categorize the tasks into groups, we also report mean and variance values for each group for ease of presentation.

\subsection{Main Results}

\input{tables/main_results}

We compare \cj with other multi-task instruction-following agents based on LLM, including Instruct GPT\!~\citep{huang2022language,instructGPT}, ReAct\!~\citep{react}, Inner Monologue\!~\citep{innermonologue}, DEPS\!~\citep{deps}. Since some methods are not originally experimented in Minecraft, we reproduce them to conform to the Minecraft specification based on prompt and feedback template design. All LLM-based methods access the LLM model through OpenAI API. And all hyper-parameters of LLM including temperature are kept as default.

The average success rates for every task group are listed in Table~\ref{tab:main_results}. \cj achieves the best performance with all meta tasks. It is important to note that in Minecraft, the technology tree can be formed by Group Wood, Stone, Iron, Gold, and Diamond. The tasks become increasingly difficult as you progress through the tree. For more difficult tasks such as obtaining a gold ingot or a diamond, the agents typically need to perform more actions and longer goal sequences in order to complete the task. As a result, the success rate of all agents decreases as the difficulty level increases. It is evident that reasoning methods (ReAct\!~\citep{react} vs. GPT\!~\citep{instructGPT,huang2022language}) and interactive re-planning with feedback (Inner Monologue\!~\citep{innermonologue} vs. GPT) effectively enhance the agent's task performance in an open world. However, these approaches still face challenges when dealing with long-horizon tasks, specifically in the Iron and Diamond group. DEPS\!~\citep{deps}, on the other hand, enables agents to accomplish diamond-related tasks through interactive long-horizon planning accompanied by descriptions and explanations. Nevertheless, its reliability remains very low at approximately 2.5\%.

In comparison to DEPS\!~\citep{deps} without memory, \cj demonstrates superior performance even in challenging tasks due to its extensive experience. In diamond-related tasks specifically, the success rate has increased by nearly 3 times (8.99\% vs 2.42\%). And \cj usually only requires 2-3 rounds of re-planning to generate the correct executable plan, whereas DEPS requires more than 6 rounds. This means that \cj saves a significant amount of LLM tokens and thinking time, enabling more efficient plan execution and providing additional steps and tokens for handling uncertainty in the environment.

Based on our observations, we have found that the bottleneck for \cj in tasks involving diamonds often lies with the Controller's inability to perfectly execute short-horizon text instructions generated by LLM. Therefore, it is worth exploring methods for generating plans that are easier for the controller to execute or improving the controller's ability to follow instructions.

\subsubsection{\cj based on different LMs}

\fancyfoot[L]{}

\begin{figure}[t]
    \centering
    \includegraphics[scale = 0.6]{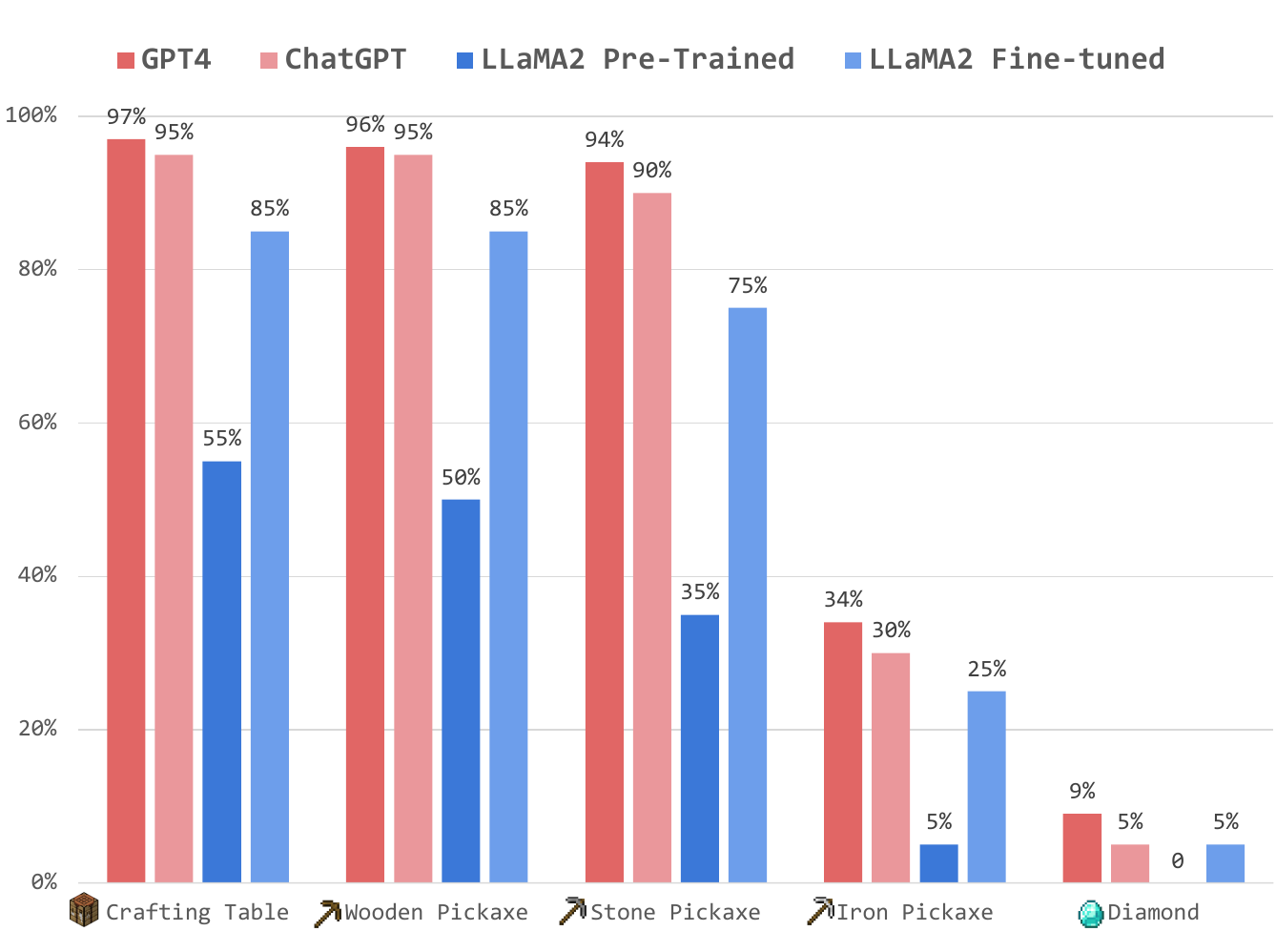}
    \caption{
        \textbf{Success rates for different language models on Minecraft tasks.} We found open-sourced LLaMA2-70B models\!~\cite{llama2} lack knowledge related to Minecraft, so the pre-trained model performs poorly. We further finetuned the LLaMA2-13B model on a Minecraft text dataset collected from the internet, and it shows performance similar to ChatGPT on Minecraft.
        }
    \label{fig:ablation_lm}
\end{figure}

We conducted ablation experiments on various Language Models, including OpenAI's ChatGPT~\cite{instructGPT} and GPT-4~\cite{gpt4}. Among these models, GPT-4 has more parameters and has been proven to outperform ChatGPT in extensive research~\cite{voyager}. We also select the open-source pre-trained LLaMA2 70B model~\cite{llama2}. Additionally, we gathered a substantial amount of Minecraft-related text from the internet as training data and further fine-tuned LLaMA2 13B. The experiments were conducted on a subset of Minecraft tasks using different language models. Each \cj learns for 4 epochs of interaction with all task sets and evaluates on task subset across at least 20 seeds. The experimental results are presented in Fig.~\ref{fig:ablation_lm}.

Table~\ref{fig:ablation_lm} demonstrates that ChatGPT, despite having fewer parameters, achieves nearly identical success rates as GPT-4. This suggests that language models equipped with memory can significantly enhance planning abilities. In Minecraft-related tasks, the open-source pre-trained LLaMA2 70B exhibits a notable performance gap compared to OpenAI models, particularly in long-horizon tasks. However, by finetuning LLaMA2 with fewer parameters, its performance on Minecraft tasks improves substantially. This indicates that the open-source model lacks knowledge specific to Minecraft and requires further finetuning for the successful completion of such tasks.

\begin{figure*}[t]
    \centering
    \includegraphics[scale = 0.6]{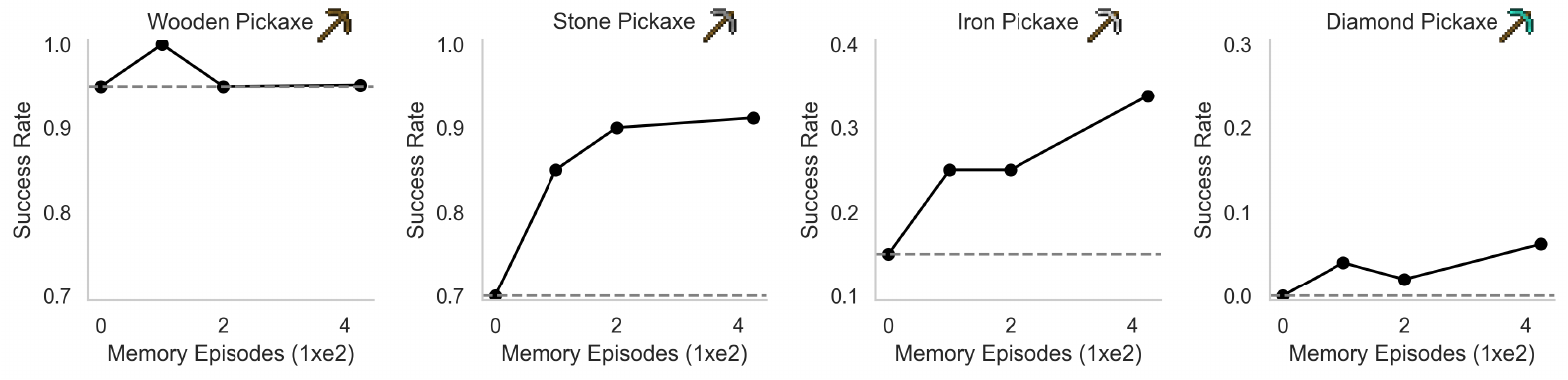}
    \caption{
        \textbf{Success rate by memory size for different items}. We evaluated the performance of \cj at different memory sizes (representing different learning stages) by measuring the success rate (\% Episodes) of completing key items on the Minecraft technology tree. As the learning progressed, we observed an improvement in completion rates for all items, with an increasing number of successful trajectories being included in memory. After 4 epochs of learning, \cj had accumulated a total of 425 successful trajectories in its memory.
        }
    \label{fig:memory_size}
\end{figure*}

\begin{figure}[t]
    \centering
    \includegraphics[scale = 0.6]{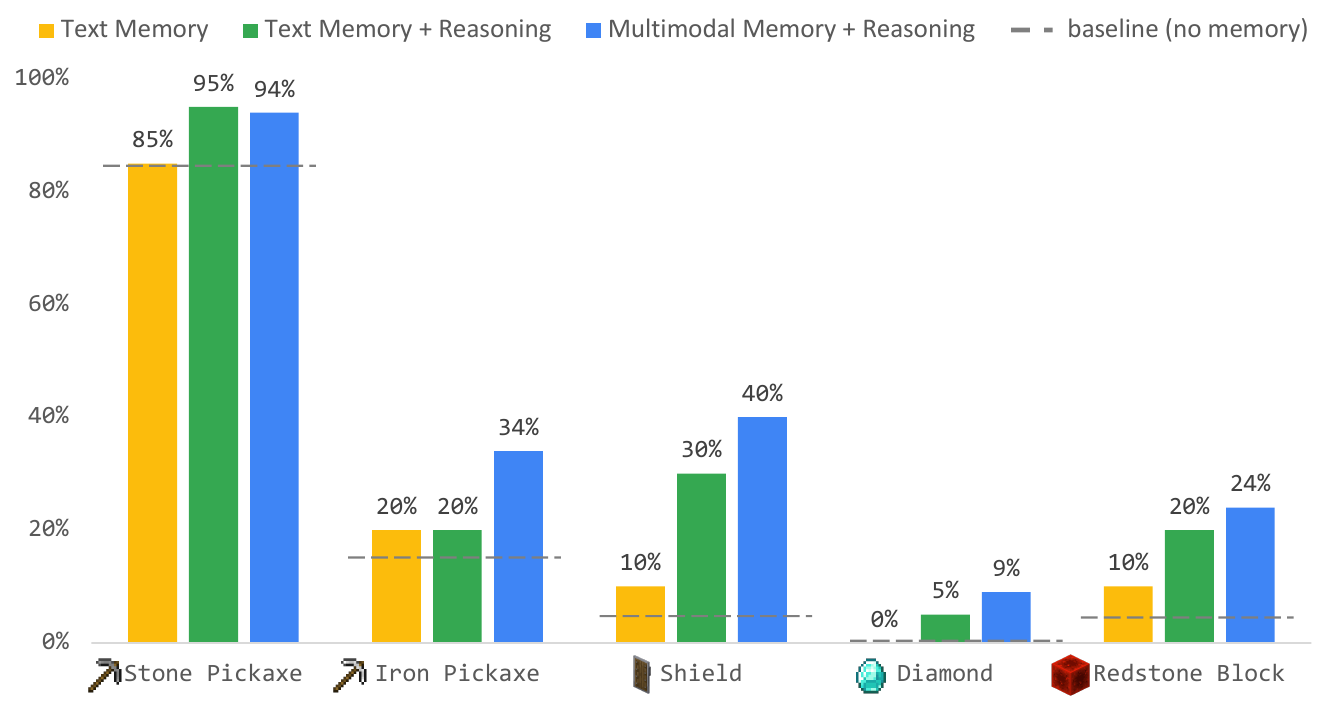}
    \caption{
        \textbf{Success rates for different retrieval methods with memory on Minecraft tasks.} \cj, which synergizes reasoning and retrieval with multimodal memory, achieves the best.
        }
    \label{fig:ablation_retrieval}
\end{figure}

\subsubsection{Ablation on Memory} \label{sec:ablation_memory}

We also conduct ablation experiments on the multimodality memory and retrieval methods. We set \cj w/o memory module as the baseline agent. We first evaluate \cj's performance with different memory sizes (representing different learning stages) as shown in Fig.~\ref{fig:memory_size}, which demonstrates the effectiveness of self-improving within \cj. We further conduct the experiments on a subset of Minecraft tasks using three different retrieval methods: retrieval with textual instruction embedding only (Text Memory), synergizing reasoning and retrieval with text embedding (Text Memory+Reasoning), and synergizing reasoning and retrieval with multimodality embedding (Multimodal Memory+Reasoning). Except for the memory and retrieval methods, all others are kept the same. The results are listed in Fig.~\ref{fig:ablation_retrieval}.

The experiments show that reasoning before retrieval can effectively improve retrieval accuracy. Retrieval based on a multimodal state including vision observation and symbolic information (e.g., inventory, location, etc) is better than only considering the text embedding.

\subsection{Long-Horizon Challenges}\label{sec:long_horizon_experiments}

\begin{figure*}[t]
    \centering
    \includegraphics[scale = 0.55]{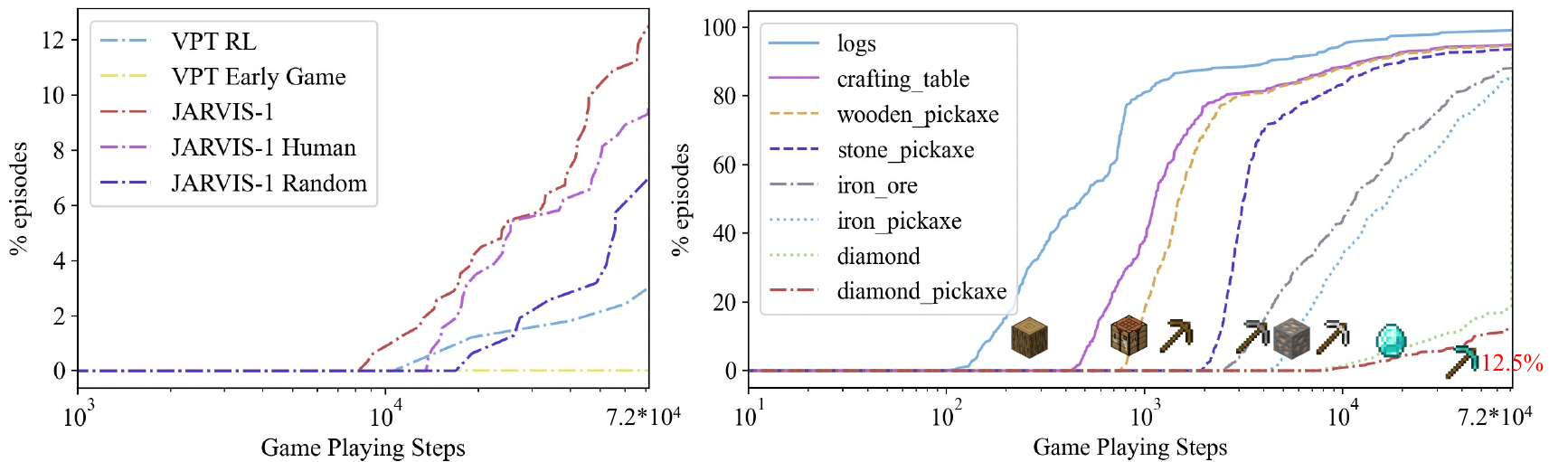}
    \caption{
        \textbf{(Left)} The success rate of different models in the \texttt{ObtainDiamondPickaxe} challenge over gameplay time. VPT RL is finetuned from VPT early game with reinforcement learning over 1.4 million episodes. \cj agent and its varients have interacted with Minecraft with over 4 epochs on all tasks in task pool. Typically, it takes a skilled person over 20 minutes (24,000 steps) to obtain a diamond pickaxe.
        \textbf{(Right)} The success rate of obtaining important intermediate items during the process of synthesizing a diamond pickaxe of \cj. This task has been evaluated over 300 times on different seeds. These curves indicate that as the game progresses, the success rates of obtaining all intermediate items are increasing, which indicates that \cj is constantly improving its skills.
        }
    \label{fig:long_horizon}
\end{figure*}

Most concurrent multi-task agents in Minecraft can only handle short-term tasks and struggle with long-horizon tasks like \texttt{CraftingDiamondPickaxe}. 
The VPT foundation model\!~\citep{vpt} is capable of accomplishing various tasks in Minecraft but lacks the ability to execute human instructions. To address this limitation, Reinforcement Learning is required to fine-tune the VPT foundation model for specific task completion. However, after fine-tuning, VPT may experience a decline in performance for other tasks while focusing on the specified task.
In contrast, Steve-1\!~\citep{steve1} has implemented goal-conditioned fine-tuning on VPT, enabling it to follow human text instructions while maintaining multitasking capabilities. However, Steve-1 primarily focuses on low-level tasks like obtaining dirt, collecting flowers, and chopping trees. When it comes to long-horizon tasks such as starting from scratch by obtaining a wooden pickaxe, Steve-1 still encounters difficulties.

DEPS\!~\citep{deps} also utilizes LLM as a planner, but it lacks the ability to learn from experience in different tasks and apply that knowledge to new ones. Additionally, DEPS is limited in its re-planning rounds due to the LM's context constraints. The experiments reveal that DEPS has a success rate of less than 50\% in generating accurate and executable plans for acquiring diamonds. The probability of DEPS successfully obtaining diamonds in the environment is approximately 0.59\%. Consequently, DEPS continues to face challenges when attempting to finish long-horizon tasks within the Minecraft world.

Even human players who have mastered the distribution pattern of diamonds achieve success rates of obtaining diamonds and crafting a diamond pickaxe (which requires at least three diamonds) within 10 minutes at approximately 15\% and 12\%, respectively. 
\cj performs better in the \texttt{ObtainDiamondPickaxe} challenge. 
Compared to the state-of-the-art model, which has undergone RL-finetuned VPT, \cj has more than doubled the success rate of obtaining a diamond pickaxe (6.22\% vs 2.5\% within 20 minutes).

To increase the chances of obtaining diamonds, we extended the game-playing time to 60 minutes (72000 game-playing steps, as shown in Figure~\ref{fig:long_horizon}). As a result, \cj's success rate in acquiring a diamond pickaxe improved from 6.2\% to 12.5\%. The graph on the right side of Figure 7 illustrates how the success rate of intermediate milestone items changes over time, indicating that \cj tends to improve with longer game-playing time. We also conduct two variants of \cj with different self-improving curricula: human-written and random-generated. All three \cj have collected experiences into memory with the curriculum for 4 epochs before evaluation in 60 minutes. The results show that \cj with a GPT-generated curriculum can finish the task within the shortest game-playing steps and achieve the best performance in 60 minutes.

In contrast, VPT's success rate barely changed when we increased the time from 20 minutes to 60 minutes (from 2.5\% to 3\%). This can be attributed to Minecraft's durability system where prolonged underground exploration often leads to pickaxe damage. When \cj's pickaxe breaks, it dynamically re-plans based on its current inventory and crafts a new one. However, VPT-RL exhibits perplexing behaviors at this stage by using inappropriate tools for mining stones or crafting unnecessary items.
This comparison demonstrates that \cj possesses superior generalization and planning abilities for long-horizon tasks.

Note that our method is designed to be multi-task in its nature and not finetuned through imitation learning on specific datasets or reinforcement learning.

\section{Related Works}

\subsection{Planning with LLM}

There have been some methods leveraging the large language model to generate action plans for high-level tasks in embodied environments~\citep{socraticmodels,reporter,mai2023llm,liu2023llm+,proagent,rladapter,jxma_llm1}.
\cite{huang2022language} decompose natural language commands into sequences of executable actions by text completion and semantic translation, while SayCan generates feasible plans for robots by jointly decoding an LLM weighted by skill affordances from value functions \citep{saycan}.
Some methods also leverage the LLM to produce the program code as plan for better executation~\citep{progprompt,codeaspolicies,text2motion}.
However, the above methods assume that the initial plan from the LLM is correct. When there are bugs in the initial plan, it's difficult for the agent to finish the task successfully. 
Recent research frequently employs LLM as an interactive planner, harnessing its self-updating capabilities to enhance the plan's executability over time~\citep{deps,reflexion,sun2023adaplanner}.
Inner Monologue~\citep{innermonologue} pilots the front of interactive planning with LLMs, which introduces the feedback (including success detection and scene description) to the planner. However, we found it could still suffer from accumulative planning errors, especially in long-horizon open-world tasks. ReAct~\citep{react} will reason about the agent state before acting, which indicates that various reasoning methods~\citep{chainofthought,treeofthought,spring} are benefitial for planning. 
LLM-based planning methods often use the fixed pretrained LLM as the agent, while we focus more on life-long and continual learning for agents in open-world environments~\citep{continual1,continual2,voyager}.
For better leveraging historical interaction between agent and environments, an explicit memory~\citep{generativeagents,gitm} for more historical chatting has been leveraged for bigger storage of agent experiences. 
However, the above methods usually rely only on a text-based environment and struggle to execute plans in partial-observed visual open-world environments.

\subsection{Minecraft Agents}
Developing generally capable agents in Minecraft to solve open-world tasks has gained increasing interests~\citep{clip4mc,minedojo,vpt,shaofei,groot,rladapter,plan4mc,gitm}.  
As an early attempt, \cite{oh2017zero} studied task generalization in a simple Minecraft environment variant. It designed a two-stage pipeline, first mastering the prerequisite skills with parameterization trick, and then learning a meta controller to execute the instructions. Moving to solve complex long-horizon tasks in Minecraft, works \citep{oh2017zero,mao2022seihai,lin2021juewu} explored the hierarchical architecture. 
In recent years, influenced by the trend of large-scale pre-training paradigms, a group of researchers have emerged, who are utilizing vast amounts of internet knowledge to train intelligent agents. 
\cite{minedojo} trained a visual-semantic alignment model, MineCLIP, using the correspondences between subtitles and video snippets available on YouTube, and used it to generate intrinsic rewards to guide policy learning.
\citep{vpt} utilizes a pre-trained inverse dynamics model to label actions in YouTube videos which are used to learn a foundation policy VPT through imitation learning. 
By bridging MineCLIP and VPT, \cite{steve1} creates a performant instruction-following policy Steve-1 to solve open-world short-horizon tasks using hindsight relabeling and unCLIP tricks. However, Steve-1 can not solve complicated process-oriented tasks due to the expressive capability of its goal space. 
\cite{groot} learns to follow reference videos as the instruction by merely watching gameplay videos, which improves the capacity of goal space and reduces the cost of policy training. 
All of these methods focus on improving the smoothness and robustness of interaction between policy and environment. Inspired by the powerful language understanding and reasoning capabilities of large language models, researchers have begun to build Minecraft agents based on LLMs. \cite{voyager} used LLM to guide the agent to explore the Minecraft world by acquiring diverse skills, making novel discoveries, and generating goal proposals. \cite{gitm} integrated LLM with text-based knowledge and memory to equip the agent with common sense and past experiences for higher reasoning efficiency. \cite{plan4mc} used LLM to guide the agent to explore the Minecraft world and interact with the environment with reinforcement learning control policies.

\section{Conclusion}
We propose a multi-task agent \cj designed for the complex environment of Minecraft, which marks a significant advancement in achieving human-like planning within an open-world setting. By leveraging pre-trained Multi-modal Language Models, \cj not only effectively interprets multimodal inputs but also adeptly translates them into actions. Its integration of a multimodal memory, which draws from both ingrained knowledge and real-time game experiences, enhances its decision-making capabilities. The empirical evidence of its prowess is evident in its impressive performance across a wide array of tasks in Minecraft. Notably, its achievement in the long-horizon diamond pickaxe task, where it achieved a completion rate that surpasses VPT by up to five times, underscores its potential and the strides made in this domain. This breakthrough sets the stage for the future of more versatile and adaptable agents in complex virtual environments.

\section*{Acknowledgments}
This work is funded in part by the National Key R\&D Program of China \#2022ZD0160301, a grant from  CCF-Tencent Rhino-Bird Open Research Fund, NSF grants \#IIS-1943641, \#IIS-1956441, \#CCF-1837129, an SRA from Meta and a research gift from Amazon Alexa AI, and a gift from RelationalAI. 
The authors sincerely thank Dr. Rita Zhang, Zhixiang Dai at NVIDIA for the valuable technical support of GPU computing.

\nocite{gong2023mindagent,mialon2023augmented}

\bibliographystyle{abbrvnat}
\nobibliography*
\bibliography{main}

\newpage

\appendix
\renewcommand\thefigure{\thesection.\arabic{figure}}
\setcounter{figure}{0}
\input{appendix}

\end{document}

%% file: config.tex
\usepackage{xcolor}
\definecolor{Gray}{gray}{0.9}
\definecolor{mygreen}{rgb}{0.0, 0.5, 0.0}
\definecolor{myred}{rgb}{0.8, 0.25, 0.33}
\definecolor{myblue}{rgb}{0.19, 0.55, 0.91}
\definecolor{uclablue}{rgb}{0.15, 0.45, 0.68}
\definecolor{boxgreen}{rgb}{0.02, 0.66, 0.02}
\definecolor{boxred}{rgb}{0.66, 0.1, 0.1}
\definecolor{boxblue}{rgb}{0.01, 0.01, 0.73}
\definecolor{mygray}{gray}{0.4}
\usepackage{times}
\usepackage{graphicx}
\usepackage{amssymb}
\usepackage{url}

\usepackage{microtype}
\usepackage{subfigure}
\usepackage{booktabs}

\usepackage{amsmath}
\usepackage{mathtools}
\usepackage{amsthm}

\usepackage{listings}
\usepackage{etoc}
\usepackage{algorithm}
\usepackage{algorithmic}
\usepackage{lipsum}
\usepackage{pifont}
\usepackage{wrapfig}
\usepackage{colortbl}
\usepackage{acronym}
\usepackage{multicol}
\usepackage{multirow}
\usepackage{makecell}
\usepackage{enumitem}
\usepackage{tabularx}
\usepackage{colortbl}
\usepackage{array}
\usepackage{xspace}
\usepackage[skip=3pt,font=small]{caption}
\usepackage{xspace}
\usepackage{mdframed}

\usepackage[export]{adjustbox}

\usepackage[utf8]{inputenc} 
\usepackage[T1]{fontenc}    
\usepackage{amsfonts}       
\usepackage{nicefrac}       

\usepackage{mathtools}
\usepackage{amssymb}
\usepackage{amsthm}

\renewcommand{\paragraph}[1]{\noindent\textbf{#1.}}

\makeatletter
\DeclareRobustCommand\onedot{\futurelet\@let@token\@onedot}
\def\@onedot{\ifx\@let@token.\else.\null\fi\xspace}
\def\eg{\emph{e.g}\onedot} 

\def\ie{\emph{i.e}\onedot}

\def\etc{\emph{etc}\onedot}

\makeatother

\acrodef{llms}[LLMs]{Large Language Models}
\acrodef{mlms}[MLMs]{Multimodal Language Models}
\newcommand{\cj}{\textsc{JARVIS}-1\xspace}

\usepackage{xcolor}


%% file: tables/mc_task_set.tex
\begin{table*}[]
\centering
\captionsetup{justification=centering}
\caption{Characteristics of 11 task groups encompassing over 200 minecraft tasks.}
\label{tab:mc_task_set}
\resizebox{0.95\linewidth}{!}{
\renewcommand\arraystretch{1.2}
\begin{tabular}{@{}llllll@{}}
\toprule
Group         & \begin{tabular}[c]{@{}l@{}}Task \\ Num.\end{tabular} & \begin{tabular}[c]{@{}l@{}}Max.\\ Steps\end{tabular} & \begin{tabular}[c]{@{}l@{}}Initial \\ Inventory\end{tabular} & Biome                                                                       & Language Instruction                                                                                    \\ \midrule
Wood          & 34                                                     & 12k                                                      & null                                                         & Plains/Forest                                                               & Pick up a \texttt{wooden\_pickaxe}.                                                                               \\
Wood-Variants & 43                                                     & 12k                                                      & null                                                         & \begin{tabular}[c]{@{}l@{}}Savanna/Jungle/Taiga\end{tabular} & Pick up a \texttt{acacia\_boat}.                                                                                  \\
Stone         & 10                                                     & 12k                                                      & iron\_axe                                                    & Plains/Forest                                                               & Craft a \texttt{furnace} given an iron axe.                                                                      \\
Iron          & 22                                                     & 12k                                                      & iron\_axe                                                    & Plains/Forest                                                               & Smelt and craft an \texttt{iron\_door} given an iron axe.                                                                           \\
Gold          & 9                                                      & 36k                                                      & iron\_axe                                                    & Plains/Forest                                                               & Smelt and craft an \texttt{golden\_axe} given an iron axe.                                                                          \\
Diamond       & 7                                                      & 36k                                                      & iron\_axe                                                    & Plains/Forest                                                               & \begin{tabular}[c]{@{}l@{}}Dig down to mine diamond and craft \texttt{diamond\_pickaxe}.\end{tabular}          \\
Redstone      & 7                                                      & 36k                                                      & iron\_axe                                                    & Plains/Forest                                                               & Mine redstone and make \texttt{dropper} given an iron axe.                                                                         \\
Blocks        & 15                                                     & 12-36k                                                   & iron\_axe                                                    & Plains/Forest                                                               & Dig down to mine \texttt{lapis\_lazuli} block.                                                                    \\
Armor         & 17                                                     & 12-36k                                                   & iron\_axe                                                    & Plains/Forest                                                               & Craft \texttt{diamond\_boots} given an iron axe and equip it.                                                                \\
Decoration    & 17                                                     & 12k                                                      & iron\_axe                                                    & Flower Forest                                                               & Obtain the \texttt{bed} and dye it \texttt{red}.                                                                          \\
Food          & 9                                                      & 12k                                                      & iron\_axe                                                    & Plains                                                                      & \begin{tabular}[c]{@{}l@{}}Kill sheep to obtain \texttt{mutton} and cook it.\end{tabular} \\ \bottomrule
\end{tabular}}
\end{table*}

%% file: tables/main_results.tex
\begin{table*}[t]
\centering
\caption{Results of \cj and baselines on Minecraft. The detailed task instructions, settings and results can be found in the Appendix.}
\label{tab:main_results}
\resizebox{0.8\linewidth}{!}{
\renewcommand\arraystretch{1.2}
\begin{tabular}{ccccccc}
\toprule
Group                                                               & Task             & GPT  & ReAct & Inner Monologue & DEPS & JARVIS-1        \\  \midrule
& \includegraphics[scale=0.04,valign=c]{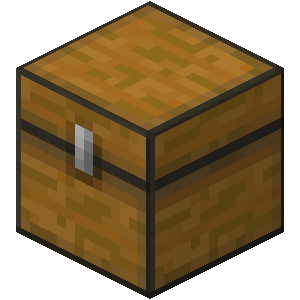}            & 26.67                               & 45.00                               & 36.67                               & 75.00                               & \textbf{91.55}       \\
\multirow{-2}{*}{Wood}                                               & AVG              & 27.30±14.86                         & 40.31±13.30                         & 60.15±19.41                         & 80.23±17.32                         & \textbf{88.84±16.82} \\ \midrule
                                                                     & \includegraphics[scale=0.08,valign=c]{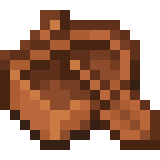}     & 6.67                                & 36.67                               & 30.00                               & 36.67                               & \textbf{60.47}       \\
\multirow{-2}{*}{\begin{tabular}[c]{@{}l@{}}Wood\\ Var\end{tabular}} & AVG              & \cellcolor[HTML]{FFFFFF}24.39±11.08 & \cellcolor[HTML]{FFFFFF}38.13±12.81 & \cellcolor[HTML]{FFFFFF}53.39±12.86 & \cellcolor[HTML]{FFFFFF}68.75±12.32 & \textbf{76.78±12.27} \\ \midrule
                                                                     & \includegraphics[scale=0.04,valign=c]{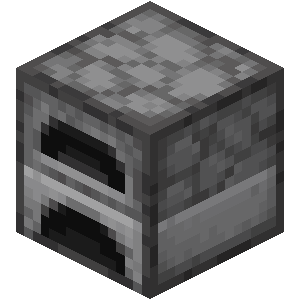}          & 20.00                               & 20.00                               & 66.67                               & 75.00                               & \textbf{94.20}       \\
\multirow{-2}{*}{Stone}                                              & AVG              & \cellcolor[HTML]{FFFFFF}20.21±12.32 & \cellcolor[HTML]{FFFFFF}39.00±12.15 & \cellcolor[HTML]{FFFFFF}52.86±16.90 & \cellcolor[HTML]{FFFFFF}69.27±7.78  & \textbf{88.69±4.87}  \\ \midrule
                                                                     & \includegraphics[scale=0.04,valign=c]{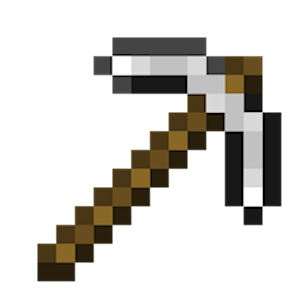}    & 0.00                                & 0.00                                & 3.33                                & 20.00                               & \textbf{33.82}       \\
                                                                     & \includegraphics[scale=0.06,valign=c]{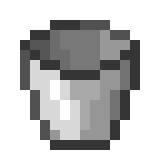}           & 3.33                                & 6.67                                & 0.00                                & 20.00                               & \textbf{38.10}       \\
\multirow{-3}{*}{Iron}                                               & AVG              & \cellcolor[HTML]{FFFFFF}3.27±2.85   & \cellcolor[HTML]{FFFFFF}4.61±3.63   & \cellcolor[HTML]{FFFFFF}5.20±5.17   & \cellcolor[HTML]{FFFFFF}16.92±4.69  & \textbf{34.63±10.61} \\ \midrule
                                                                     & \includegraphics[scale=0.08,valign=c]{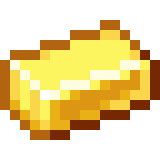}      & 0.00                                & 2.00                                & 2.00                                & 6.00                                & \textbf{14.49}       \\
\multirow{-2}{*}{Gold}                                               & AVG              & \cellcolor[HTML]{FFFFFF}0.00±0.00   & \cellcolor[HTML]{FFFFFF}0.45±0.60   & \cellcolor[HTML]{FFFFFF}0.59±0.64   & \cellcolor[HTML]{FFFFFF}2.20±1.55   & \textbf{6.85±4.71}   \\ \midrule
                                                                     & \includegraphics[scale=0.04,valign=c]{minecraft/diamond.png}          & 0.00                                & 0.00                                & 1.00                                & 2.00                                & \textbf{9.20}        \\
                                                                     & \includegraphics[scale=0.08,valign=c]{minecraft/diamond_pickaxe.png} & 0.00                                & 0.00                                & 0.00                                & 2.50                                & \textbf{6.22}        \\
\multirow{-3}{*}{Diamond}                                            & AVG              & \cellcolor[HTML]{FFFFFF}0.00±0.00   & \cellcolor[HTML]{FFFFFF}0.35±0.48   & \cellcolor[HTML]{FFFFFF}0.96±0.67   & \cellcolor[HTML]{FFFFFF}2.42±1.01   & \textbf{8.99±2.68}   \\ \midrule
                                                                     & \includegraphics[scale=0.04,valign=c]{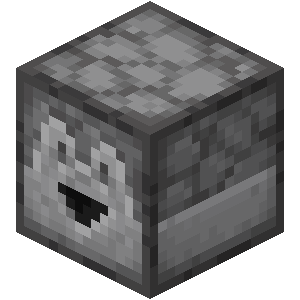}          & 0.00                                & 2.00                                & 0.00                                & 10.00                               & \textbf{22.78}       \\
\multirow{-2}{*}{Redstone}                                           & AVG              & \cellcolor[HTML]{FFFFFF}1.04±1.30   & \cellcolor[HTML]{FFFFFF}1.14±1.18   & \cellcolor[HTML]{FFFFFF}0.69±1.68   & \cellcolor[HTML]{FFFFFF}6.02±3.61   & \textbf{17.51±9.34}  \\ \midrule
                                                                     & \includegraphics[scale=0.08,valign=c]{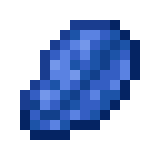}    & 16.67                               & 33.33                               & 43.33                               & 53.33                               & \textbf{86.67}       \\
\multirow{-2}{*}{Blocks}                                             & AVG              & \cellcolor[HTML]{FFFFFF}45.64±33.88 & \cellcolor[HTML]{FFFFFF}49.35±30.51 & \cellcolor[HTML]{FFFFFF}55.71±29.43 & \cellcolor[HTML]{FFFFFF}58.02±27.68 & \textbf{80.34±21.09} \\ \midrule
                                                                     & \includegraphics[scale=0.08,valign=c]{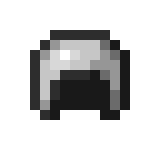}     & 6.67                                & 0.00                                & 10.00                               & 10.00                               & \textbf{30.30}       \\
\multirow{-2}{*}{Armor}                                              & AVG              & \cellcolor[HTML]{FFFFFF}1.36±2.25   & \cellcolor[HTML]{FFFFFF}0.50±0.88   & \cellcolor[HTML]{FFFFFF}3.10±4.71   & \cellcolor[HTML]{FFFFFF}3.71±3.78   & \textbf{13.44±14.62} \\ \midrule
                                                                     & \includegraphics[scale=0.04,valign=c]{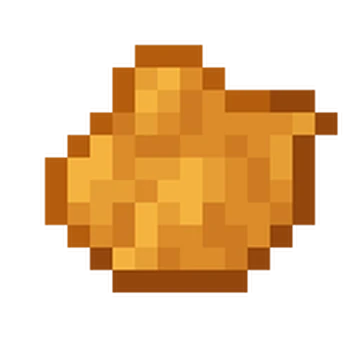}     & 15.00                               & 15.00                               & 15.00                               & 25.00                               & \textbf{50.00}       \\
\multirow{-2}{*}{Decoration}                                         & AVG              & \cellcolor[HTML]{FFFFFF}17.12±11.59 & \cellcolor[HTML]{FFFFFF}17.13±9.19  & \cellcolor[HTML]{FFFFFF}12.03±10.19 & \cellcolor[HTML]{FFFFFF}29.59±15.94 & \textbf{46.67±23.39} \\ \midrule
                                                                     & \includegraphics[scale=0.08,valign=c]{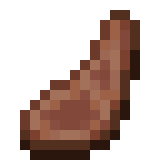}   & 13.33                               & 16.67                               & 25.00                               & 16.67                               & \textbf{43.55}       \\
\multirow{-2}{*}{Food}                                               & AVG              & \cellcolor[HTML]{FFFFFF}9.40±4.29   & \cellcolor[HTML]{FFFFFF}15.56±6.83  & \cellcolor[HTML]{FFFFFF}20.78±11.99 & \cellcolor[HTML]{FFFFFF}22.85±8.15  & \textbf{46.75±11.16} \\ \bottomrule
\end{tabular}}
\end{table*}

%% file: appendix.tex




\section{Implementation Details}\label{sec:implementation_details}

\subsection{Controller}\label{sec:controller}

Tasks in Minecraft are usually related to \texttt{mine} and \texttt{craft} goals. The \texttt{mine} goals require the agent to collect raw materials from the environment using the appropriate tools. The \texttt{craft} goals ask the agent to use the recipe to generate new items with existing materials in inventory. 
The \texttt{mine} goals are achieved through STEVE-1~\citep{steve1} with text condition during implementation. The environment can directly executes the \texttt{craft} and \texttt{smelt} actions (\texttt{craft}/\texttt{smelt} with argument), which are same as MineDojo~\citep{minedojo} .

\subsection{Interactive Planner}

\cj relies on the Multi-modal Language Model for planning, self-checking, and self-explaining, and can accept three types of inputs: visual images, language, and symbolic information (including inventory, located position, home, current life statistics, etc.). Specifically, this is a hybrid model with language processing capabilities derived from the GPT model~\citep{gpt4}. The visual ability comes from MineCLIP~\citep{minedojo}. We collected approximately 1000 Minecraft text data from the internet and calculated the similarity between the current vision observation and these text data. Text above the similarity threshold will be selected into the GPT model's prompt. Symbolic information is converted into natural language text through a designed template. All modalities are ultimately captured as language and processed by the GPT model.

Different modules in \cj (e.g. self-check and self-explain) are completed through MLM based on different prompts. The specific prompt design are shown below.

\input{appendix/planning}

\input{appendix/parser}

\input{appendix/explanation}

\input{appendix/self_check}



\subsection{Memory}

Our memory records every successful trajectory experience of \cj, including the task goals that the agent needs to execute, the actual goal sequence (plan) executed by the agent, and the state (visual observation and symbolic information returned from the environment) when the agent completes the task. In specific implementation, memory is a list where each trajectory experience is encoded as a dictionary, including the keys \texttt{task}, \texttt{state}, and \texttt{plan}.

\section{Environment Setting}

Our Minecraft environment is a hybrid between MineRL\!~\citep{minerl} and the MCP-Reborn (github.com/Hexeption/MCP-Reborn) Minecraft modding package. Unlike the regular Minecraft game, in which the server (or the "world") always runs at 20Hz and the client runs as fast as rendering

\subsection{Observation Space}

The environmental observations consist of two parts. The first part is the raw pixels from the Minecraft game that players would see, including overlays such as the hotbar, health indicators, and animations of a moving hand in response to attack or "use" actions. The field of view, GUI scale, and brightness parameters are consistent with VPT~\citep{vpt}. 
The second part includes auxiliary information about the agent's current environment, such as its location and weather conditions. Human players can obtain this information by pressing F3. The specific observation details we include are shown in Table~\ref{tab:observation_space}.

\input{tables/observation_space}

Note that no high-level observations like voxels and lidar information in Minedojo\!~\citep{minedojo} can be accessed by agents.
During the actual inference process, the controller only perceives the raw pixels and interacts with the environment, which is the same with VPT\!~\citep{vpt} models. The agent will access information from the environment to generate the text condition of the controller.

\subsection{Action Space}
We design a hybrid action space. Some are directly available to human players, including keypresses, mouse movements, and clicks, which are similar to MineRL v1.0~\citep{minerl} used by VPT~\citep{vpt}. The keypresses and clicks are binary functional actions, including \texttt{forward}, \texttt{jump}, \texttt{use} and \texttt{attack} etc. In addition to the binary (on/off) keypress actions, our action space also includes mouse movements. When the in-game GUIs (press "E" to open inventory) are closed, the mouse's X and Y actions control the agent's yaw and pitch. However, when the GUI is open, camera actions move the mouse cursor on the screen. In Minecraft, precise mouse movements are needed to interact with the inventory for tasks such as crafting and smelting. On the other hand, mining and navigating the world can be done using broader mouse actions. To be enable to achieve both the same action space, we abstract the \texttt{craft} and \texttt{smelt} action with GUI into functional binary actions, which are same as MineDojo~\citep{minedojo}. The detailed action space are described in Table~\ref{tab:action_space}.

\input{tables/action_space}

\subsection{Rules}

We choose to conduct the test in survival mode of Minecraft 1.16.5. For each environment reset, we have added the following rules:
\begin{itemize}
    \item \texttt{/difficulty peaceful}: Set the difficulty of the environment to peaceful mode.
    \item \texttt{/gamerule doDaylightCycle false}: Set the environment to daytime forever.
    \item \texttt{/gamerule keepInventory true}: Set agent to not drop items upon death. We have added a time limit for each task, within which if the player dies, they will respawn at the spawn point and retain their previous inventory contents.
    \item \texttt{/effect give @a night\_vision 99999 250 true}: In order to facilitate the display of agent behavior, we have added night vision effects to the agent.
\end{itemize}










\section{Results and Details of 200+ tasks in Minecraft Universe Benchmark}\label{sec:detailed_results}

We list the evaluation task set belows with details including task name, maximum steps, initial inventory, biome, and language instructions.
We also show the evaluation times across different seeds and successful episodes rate.
Note that all tasks are evaluated in Minecraft 1.16.5 Survival Mode.

\input{tables/results_wood}


\input{tables/results_stone}

\input{tables/results_iron}

\input{tables/results_gold}

\input{tables/results_diamond}

\input{tables/results_redstone}

\input{tables/results_blocks}

\input{tables/results_armor}

\input{tables/results_decoration}

\input{tables/results_food}






%% file: appendix/planning.tex
\definecolor{codeblue}{rgb}{0.25,0.5,0.5}
\definecolor{codekw}{rgb}{0.85, 0.18, 0.50}
\definecolor{keywordgreen}{rgb}{0,0.6,0}
\lstset{
  backgroundcolor=\color{gray!10},
  basicstyle=\fontsize{8pt}{9pt}\ttfamily\selectfont,
  columns=fullflexible,
  breaklines=true,
  captionpos=b,
  commentstyle=\fontsize{7.5pt}{7.5pt}\color{codeblue},
  keywords = {User, Assistant, System}, 
  keywordstyle = {\textbf},
  caption={Planning prompt in \cj},
  label={lst:planning_prompt}
}
\begin{lstlisting} % [language=python]  

System: You are a helper agent in Minecraft. You need to generate the sequences of goals for a certain task in Minecraft. Just refer the history dialogue to give the plan consist of template. Do not explain or give any other instruction. 

==========
User: My current inventory has nothing. I current locate in plains. How to obtain 1 wooden_pickaxe in Minecraft step-by-step?
Assistant: The code for obtaining 1 wooden_pickaxe is as follows:
def craft_wooden_axe(initial_inventory={}):
    mine(obj = {"log":3}, tool = None)  # step 1: chop down trees to mine logs
    craft(obj = {"planks":12}, materials = {"log":3}, tool = None)   # step 2: craft 12 planks from 3 logs
    craft(obj = {"stick":4}, materials = {"planks":2}, tool = None) # step 3: craft 4 sticks from 2 planks
    craft(obj = {"crafting_table":1}, materials = {"planks":4}, tool = None) # step 4: craft 1 crafting_table from 4 planks
    craft(obj = {"wooden_axe":1}, {"planks": 3, "stick": 2}, "crafting_table") # step 5: craft 1 wooden_axe from 3 planks and 2 sticks on crafting table
    return "wooden_axe"
User: [Description] I succeed in step 1, 2, 3, 4, 5. 
I finish all steps and I obtain 1 wooden_axe successfully.

==========
User: My current inventory has <inventory>. <visual observation>. How to obtain 1 stone_sword in Minecraft step-by-step?
Assistant: 

\end{lstlisting}

%% file: appendix/parser.tex
\definecolor{codeblue}{rgb}{0.25,0.5,0.5}
\definecolor{codekw}{rgb}{0.85, 0.18, 0.50}
\definecolor{keywordgreen}{rgb}{0,0.6,0}
\lstset{
  backgroundcolor=\color{gray!10},
  basicstyle=\fontsize{8pt}{8.5pt}\ttfamily\selectfont,
  columns=fullflexible,
  breaklines=true,
  captionpos=b,
  commentstyle=\fontsize{7.5pt}{7.5pt}\color{codeblue},
  keywords = {User, Assistant, System}, 
  keywordstyle = {\textbf},
  caption={Goal parsing prompt in \cj},
  label={lst:parser_prompt}
}

\begin{lstlisting} % [language=python] 

System: Extract the action name, action type, goal object, tool and action rank from the input text.

User: mine({"log":3}, null); # step 1: chop down trees to mine logs
Assistant:
name: mine_log
text condition: chop down trees to mine logs
action: mine
object_item: log
object_number: 3
tool: null
rank: 1
###

input: craft({"planks":12}, {"log":3}, null); # step 2: craft 12 planks from 3 log
Assistant:
name: craft_planks
condition: craft 12 planks from 3 log
action: craft
object_item: planks
object_number: 12
materials: {"log":3}
tool: null
rank: 2
###

\end{lstlisting}

%% file: appendix/explanation.tex
\definecolor{codeblue}{rgb}{0.25,0.5,0.5}
\definecolor{codekw}{rgb}{0.85, 0.18, 0.50}
\definecolor{keywordgreen}{rgb}{0,0.6,0}
\lstset{
  backgroundcolor=\color{gray!10},
  basicstyle=\fontsize{8pt}{8.5pt}\ttfamily\selectfont,
  columns=fullflexible,
  breaklines=true,
  captionpos=b,
  commentstyle=\fontsize{7.5pt}{7.5pt}\color{codeblue},
  keywords = {System, User, Assistant}, 
  keywordstyle = {\textbf},
  caption={Self-explaining prompt in \cj},
  label={lst:explainer_prompt}
}

\begin{lstlisting} % [language=python] 

System: Here are some actions that the agent fails to perform in Minecraft. Please give the explanation of action execution failure according to the current inventory information of the agent.

###
User:
Failed Action: mine({"iron_ore":1}, null); # step 5: dig down to mine iron blocks
Current Inventory: null
Assistant: Because mining iron_ore needs to use the tool stone_pickaxe, but my inventory does not have stone_pickaxe. So I need to craft stone_pickaxe first.

###
User:
Failed Action: craft({"stone_pickaxe":1}, {"cobblestone":3, "stick":2}, "crafting_table"); # step 1: craft 1 stone_pickaxe from 3 cobblestone and 2 stick, on crafting_table
Current Inventory: null
Assistant: Because crafting stone_pickaxe needs to have 3 cobblestone and 2 stick in inventory, but my inventory does not have cobblestone and stick. So I need to mine cobblestone and craft stick first.

### 
User:
Failed Action: craft({"stick":4}, {"planks":2}, null); # step 3: craft 4 stick from 2 planks first
Current Inventory: null
Assistant: Because crafting stick needs to have planks in inventory, but my inventory does not have planks. So I need to craft planks first.

###

\end{lstlisting}

%% file: appendix/self_check.tex
\definecolor{codeblue}{rgb}{0.25,0.5,0.5}
\definecolor{codekw}{rgb}{0.85, 0.18, 0.50}
\definecolor{keywordgreen}{rgb}{0,0.6,0}
\lstset{
  backgroundcolor=\color{gray!10},
  basicstyle=\fontsize{8pt}{9pt}\ttfamily\selectfont,
  columns=fullflexible,
  breaklines=true,
  captionpos=b,
  commentstyle=\fontsize{7.5pt}{7.5pt}\color{codeblue},
  keywords = {User, Assistant, System}, 
  keywordstyle = {\textbf},
  caption={Self-checking prompt in \cj},
  label={lst:self_check_prompt}
}
\begin{lstlisting} % [language=python]  

System: You are a helper agent in Minecraft. Check the plan whether can be finished.

==========
User: My current inventory has nothing. I current locate in plains. My task is to obtain 1 wooden axe in Minecraft step-by-step. This is my plan:
def craft_wooden_axe(initial_inventory={}):
    mine(obj = {"log":3}, tool = None)  # step 1: chop down trees to mine logs
    craft(obj = {"planks":12}, materials = {"log":3}, tool = None)   # step 2: craft 12 planks from 3 logs
    craft(obj = {"stick":4}, materials = {"planks":2}, tool = None) # step 3: craft 4 sticks from 2 planks
    craft(obj = {"wooden_axe":1}, {"planks": 3, "stick": 2}, null) # step 4: craft 1 wooden_axe from 3 planks and 2 sticks
    return "wooden_axe"

Assistant:
My current inventory is empty.
1. I want to mine 3 logs. Mining logs do not require tools. My inventory will has 3 logs.
2. I want to craft 12 planks from 3 logs. 1 log can produce 4 planks. My inventory will add 12 planks, and delete 3 logs. My inventory will has 12 planks.
3. I want to craft 4 stick from 2 planks. 2 planks can produce 4 planks. My inventory will add 4 stick and delete 2 planks. My inventory will has 10 planks, 4 stick.
4. I want to craft 1 wooden pickaxe from 3 planks and 2 sticks. Crafting wooden pickaxe requires crafting_table. But I do not have crafting_table in inventory. This action will failed. 
Return: Step 4 will failed because of lacking of crafting_table.

\end{lstlisting}

%% file: tables/observation_space.tex
\begin{table}[H]
\centering
\captionsetup{justification=centering}
\caption{The observation space we use in Minecraft.}
\label{tab:observation_space}
\resizebox{0.8\linewidth}{!}{
\renewcommand\arraystretch{1.1}
\begin{tabular}{@{}lll@{}}
\toprule
Sources         & Shape       & Description                                                                                                                                                                                                                                                                                                                                \\ \midrule
pov             & (640, 360, 3) & Ego-centric RGB frames.                                                                                                                                                                                                                                                                                                                    \\ \midrule
player\_pos     & (5,)        & The coordinates of (x,y,z), pitch, and yaw of the agent.                                                                                                                                                                                                                                                                                   \\ \midrule
location\_stats & (9,)        & \begin{tabular}[c]{@{}l@{}}The environmental information of the agent's current position, \\ including \texttt{biome\_id}, \texttt{sea\_level}, \texttt{can\_see\_sky}, \texttt{is\_raining} etc.\end{tabular}                                                                                                                                                                 \\ \midrule
inventory       & (36,)       & \begin{tabular}[c]{@{}l@{}}The items in the current inventory of the agent, including \\ the \texttt{type} and corresponding \texttt{quantity} of each item in each slot. \\ If there is no item, it will be displayed as \texttt{air}.\end{tabular}                                                                                                                \\ \midrule
equipped\_items & (6,)        & \begin{tabular}[c]{@{}l@{}}The current equipment of the agent, including \texttt{mainhand}, \texttt{offhand}, \\ \texttt{chest}, \texttt{feet}, \texttt{head}, and \texttt{legs} slots. Each slot contains \texttt{type}, \texttt{damage}, \\ and \texttt{max\_damage} information.\end{tabular}                                                                                                                           \\ \midrule
event\_info     & (5,)        & \begin{tabular}[c]{@{}l@{}}The events that occur in the current step of the game, including \\ \texttt{pick\_up} (picking up items), \texttt{break\_item} (breaking items), \\ \texttt{craft\_item} (crafting items using a crafting table or crafting grid), \\ \texttt{mine\_block} (mining blocks by suitable tools), and \\ \texttt{kill\_entity} (killing game mobs).\end{tabular} \\ \bottomrule
\end{tabular}}
\end{table}

%% file: tables/action_space.tex
\begin{table}[H]
\centering
\captionsetup{justification=centering}
\caption{The action space we use in Minecraft.}
\label{tab:action_space}
\resizebox{0.8\linewidth}{!}{
\renewcommand\arraystretch{1.1}
\begin{tabular}{@{}cccl@{}}
\toprule
\textbf{Index} & \textbf{Action}  & \textbf{Human Action} & \textbf{Description}                                                                                                                                                                                                                                                                                             \\ \midrule
1              & Forward          & key W                 & Move forward.                                                                                                                                                                                                                                                                                                    \\
2              & Back             & key S                 & Move backward.                                                                                                                                                                                                                                                                                                   \\
3              & Left             & key A                 & Strafe left.                                                                                                                                                                                                                                                                                                     \\
4              & Right            & key D                 & Strafe right.                                                                                                                                                                                                                                                                                                    \\
5              & Jump             & key Space             & Jump. When swimming, keeps the player afloat.                                                                                                                                                                                                                                                                    \\
6              & Sneak            & key left Shift        & Slowly move in the current direction of movement. \\
7              & Sprint           & key left Ctrl         & Move quickly in the direction of current motion.                                                                                                                                                                                                                                                                 \\
8              & Attack           & left Button     & Destroy blocks (hold down); Attack entity (click once).                                                                                                                                                                                                                                                          \\
9              & Use              & right Button    & Interact with the block that the player is currently looking at.                                                                                                 \\
10             & hotbar.{[}1-9{]} & keys 1 - 9            & Selects the appropriate hotbar item.                                                                                                                                                \\
11             & Yaw              & move Mouse X      & Turning; aiming; camera movement.Ranging from -180 to +180.                                                                                                                                                                                                                                                      \\
12             & Pitch            & move Mouse Y      & Turning; aiming; camera movement.Ranging from -180 to +180.  \\

13             & Equip            & -      & Equip the item in main hand from inventory.  \\

14             & Craft            & -      & Execute a crafting recipe to obtain new item.  \\

15             & Smelt            & -      & Execute a smelting recipe to obtain new item.

\\ \bottomrule
\end{tabular}}
\end{table}

%% file: tables/results_wood.tex
\begin{table}[]
\centering
\captionsetup{justification=centering}
\caption{The results of our agent on various tasks in the \texttt{Wood} group.}
\label{tab:wood_results}
\resizebox{0.9\linewidth}{!}{
\renewcommand\arraystretch{1.1}
\begin{tabular}{@{}lllllll@{}}
\toprule
Task                         & \begin{tabular}[c]{@{}l@{}}Max.\\ Steps\end{tabular} & \begin{tabular}[c]{@{}l@{}}Initial \\ Inventory\end{tabular} & Biome         & \begin{tabular}[c]{@{}l@{}}Success \\ Rate\end{tabular} & \begin{tabular}[c]{@{}l@{}}Eval\\ Times\end{tabular} & Language Instruction                     \\ \midrule
wooden\_shovel               & 12000                                                & null                                                         & Plains/Forest & 0.9028                                                  & 72                                                   & Pick up a wooden\_shovel given nothing.  \\
wooden\_pickaxe              & 12000                                                & null                                                         & Plains/Forest & 0.9516                                                  & 62                                                   & Pick up a wooden\_pickaxe given nothing. \\
wooden\_axe                  & 12000                                                & null                                                         & Plains/Forest & 0.8909                                                  & 55                                                   & Pick up a wooden\_axe given nothing.     \\
wooden\_hoe                  & 12000                                                & null                                                         & Plains/Forest & 0.9318                                                  & 44                                                   & Pick up a wooden\_hoe given nothing.     \\
stick                        & 12000                                                & null                                                         & Plains/Forest & 1                                                       & 86                                                   & Pick up a stick given nothing.           \\
wooden\_sword                & 12000                                                & null                                                         & Plains/Forest & 0.9242                                                  & 66                                                   & Pick up a wooden\_sword given nothing.   \\
composter                    & 12000                                                & null                                                         & Plains/Forest & 0.7872                                                  & 47                                                   & Pick up a composter given nothing.       \\
barrel                       & 12000                                                & null                                                         & Plains/Forest & 0.7544                                                  & 57                                                   & Pick up a barrel given nothing.          \\
crafting\_table              & 12000                                                & null                                                         & Plains/Forest & 0.9706                                                  & 68                                                   & Pick up a crafting\_table given nothing. \\
{\color[HTML]{000000} chest} & 12000                                                & null                                                         & Plains/Forest & 0.9155                                                  & 71                                                   & Pick up a chest given nothing.           \\
ladder                       & 12000                                                & null                                                         & Plains/Forest & 0.9737                                                  & 76                                                   & Pick up a ladder given nothing.          \\
bowl                         & 12000                                                & null                                                         & Plains/Forest & 0.9149                                                  & 47                                                   & Pick up a bowl given nothing.            \\ \midrule
oak\_wood                    & 12000                                                & null                                                         & Forest        & 0.9868                                                  & 76                                                   & Pick up a oak\_wood in Forest.           \\
oak\_slab                    & 12000                                                & null                                                         & Forest        & 0.9506                                                  & 81                                                   & Pick up a oak\_slab in Forest.           \\
oak\_planks                  & 12000                                                & null                                                         & Forest        & 0.9659                                                  & 88                                                   & Pick up a oak\_planks in Forest.         \\
oak\_log                     & 12000                                                & null                                                         & Forest        & 1                                                       & 65                                                   & Pick up a oak\_log in Forest.            \\
oak\_button                  & 12000                                                & null                                                         & Forest        & 0.9153                                                  & 59                                                   & Pick up a oak\_button in Forest.         \\
oak\_door                    & 12000                                                & null                                                         & Forest        & 0.8732                                                  & 71                                                   & Pick up a oak\_door in Forest.           \\
oak\_fence                   & 12000                                                & null                                                         & Forest        & 0.8                                                     & 60                                                   & Pick up a oak\_fence in Forest.          \\
oak\_fence\_gate             & 12000                                                & null                                                         & Forest        & 0.9322                                                  & 59                                                   & Pick up a oak\_fence\_gate in Forest.    \\
oak\_trapdoor                & 12000                                                & null                                                         & Forest        & 0.8861                                                  & 79                                                   & Pick up a oak\_trapdoor in Forest.       \\
oak\_boat                    & 12000                                                & null                                                         & Forest        & 0.9074                                                  & 54                                                   & Pick up a oak\_boat in Forest.           \\
oak\_sign                    & 12000                                                & null                                                         & Forest        & 0.9                                                       & 40                                                    & Pick up a oak\_sign in Forest.           \\ \midrule
birch\_wood                  & 12000                                                & null                                                         & Forest        & 0.9474                                                  & 57                                                   & Pick up a birch\_wood in Forest.         \\
birch\_slab                  & 12000                                                & null                                                         & Forest        & 0.9231                                                  & 65                                                   & Pick up a birch\_slab in Forest.         \\
birch\_planks                & 12000                                                & null                                                         & Forest        & 0.9714                                                  & 70                                                   & Pick up a birch\_planks in Forest.       \\
birch\_log                   & 12000                                                & null                                                         & Forest        & 0.9833                                                  & 60                                                   & Pick up a birch\_log in Forest.          \\
birch\_button                & 12000                                                & null                                                         & Forest        & 0.9245                                                  & 53                                                   & Pick up a birch\_button in Forest.       \\
birch\_door                  & 12000                                                & null                                                         & Forest        & 0.8431                                                  & 51                                                   & Pick up a birch\_door in Forest.         \\
birch\_fence                 & 12000                                                & null                                                         & Forest        & 0.8                                                     & 30                                                   & Pick up a birch\_fence in Forest.        \\
birch\_fence\_gate           & 12000                                                & null                                                         & Forest        & 0.9355                                                  & 62                                                   & Pick up a birch\_fence\_gate in Forest.  \\
birch\_trapdoor              & 12000                                                & null                                                         & Forest        & 0.9524                                                  & 63                                                   & Pick up a birch\_trapdoor in Forest.     \\
birch\_boat                  & 12000                                                & null                                                         & Forest        & 0.8906                                                  & 64                                                   & Pick up a birch\_boat in Forest.         \\
birch\_sign                  & 12000                                                & null                                                         & Forest        & 0.9                                                     & 60                                                   & Pick up a birch\_sign in Forest.         \\ \bottomrule
\end{tabular}}
\end{table}

%% file: tables/results_stone.tex
\begin{table}[]
\centering
\captionsetup{justification=centering}
\caption{The results of our agent on various tasks in the \texttt{Stone} group.}
\label{tab:stone_results}
\resizebox{0.9\linewidth}{!}{
\renewcommand\arraystretch{1.1}
\begin{tabular}{@{}lllllll@{}}
\toprule
Task                         & \begin{tabular}[c]{@{}l@{}}Max.\\ Steps\end{tabular} & \begin{tabular}[c]{@{}l@{}}Initial \\ Inventory\end{tabular} & Biome         & \begin{tabular}[c]{@{}l@{}}Success \\ Rate\end{tabular} & \begin{tabular}[c]{@{}l@{}}Eval\\ Times\end{tabular} & Language Instruction                       \\ \midrule
stone\_shovel                & 12000                                                & iron\_axe                                                    & Plains/Forest & 0.8514                                                  & 74                                                   & Craft a stone\_shovel given an iron\_axe.  \\
stone\_pickaxe               & 12000                                                & iron\_axe                                                    & Plains/Forest & 0.9118                                                  & 68                                                   & Craft a stone\_pickaxe given an iron\_axe. \\
stone\_axe                   & 12000                                                & iron\_axe                                                    & Plains/Forest & 0.9123                                                  & 57                                                   & Craft a stone\_axe given an iron\_axe.     \\
stone\_hoe                   & 12000                                                & iron\_axe                                                    & Plains/Forest & 0.9459                                                  & 74                                                   & Craft a stone\_hoe given an iron\_axe.     \\
stone                        & 12000                                                & iron\_axe                                                    & Plains/Forest & 0.8413                                                  & 63                                                   & Craft a stone given an iron\_axe.          \\
charcoal                     & 12000                                                & iron\_axe                                                    & Plains/Forest & 0.8947                                                  & 76                                                   & Craft a charcoal given an iron\_axe.       \\
smoker                       & 12000                                                & iron\_axe                                                    & Plains/Forest & 0.7867                                                  & 75                                                   & Craft a smoker given an iron\_axe.         \\
stone\_sword                 & 12000                                                & iron\_axe                                                    & Plains/Forest & 0.8831                                                  & 77                                                   & Craft a stone\_sword given an iron\_axe.   \\
furnace                      & 12000                                                & iron\_axe                                                    & Plains/Forest & 0.942                                                   & 69                                                   & Craft a furnace given an iron\_axe.        \\
torch & 12000                                                & iron\_axe                                                    & Plains/Forest & 0.9                                                     & 30                                                   & Craft a torch given an iron\_axe.          \\ \bottomrule
\end{tabular}
}
\end{table}

%% file: tables/results_iron.tex
\begin{table}[]
\centering
\captionsetup{justification=centering}
\caption{The results of our agent on various tasks in the \texttt{Iron} group.}
\label{tab:iron_results}
\resizebox{0.9\linewidth}{!}{
\renewcommand\arraystretch{1.1}
\begin{tabular}{@{}lllllll@{}}
\toprule
Task                             & \begin{tabular}[c]{@{}l@{}}Max.\\ Steps\end{tabular} & \begin{tabular}[c]{@{}l@{}}Initial \\ Inventory\end{tabular} & Biome         & \begin{tabular}[c]{@{}l@{}}Success \\ Rate\end{tabular} & \begin{tabular}[c]{@{}l@{}}Eval\\ Times\end{tabular} & Language Instruction                                                                           \\ \midrule
iron\_axe                        & 12000                                                & null                                                         & Plains/Forest & 0.3333                                                  & 60                                                   & Smelt and craft an iron\_axe.                                                                  \\
iron\_pickaxe                    & 12000                                                & iron\_axe                                                    & Plains/Forest & 0.3382                                                  & 68                                                   & Smelt and craft an iron\_pickaxe.                                                              \\
iron\_shovel                     & 12000                                                & iron\_axe                                                    & Plains/Forest & 0.338                                                   & 71                                                   & Smelt and craft an iron\_shovel.                                                               \\
iron\_sword                      & 12000                                                & iron\_axe                                                    & Plains/Forest & 0.3288                                                  & 73                                                   & Smelt and craft an iron\_sword.                                                                \\
iron\_trapdoor                   & 12000                                                & iron\_axe                                                    & Plains/Forest & 0.3151                                                  & 73                                                   & Smelt and craft an iron\_trapdoor.                                                             \\
iron\_door                       & 12000                                                & iron\_axe                                                    & Plains/Forest & 0.2836                                                  & 67                                                   & Smelt and craft an iron\_door.                                                                 \\
iron\_ingot                      & 12000                                                & iron\_axe                                                    & Plains/Forest & 0.5479                                                  & 73                                                   & Smelt and craft an iron\_ingot.                                                                \\
bucket                           & 12000                                                & iron\_axe                                                    & Plains/Forest & 0.381                                                   & 42                                                   & Smelt and craft a bucket.                                                                      \\
rail                             & 12000                                                & iron\_axe                                                    & Plains/Forest & 0.3226                                                  & 62                                                   & Smelt and craft a rail.                                                                        \\
{\color[HTML]{000000} minecart}  & 12000                                                & iron\_axe                                                    & Plains/Forest & 0.2833                                                  & 60                                                   & Smelt and craft a minecart.                                                                    \\ 
smithing\_table                  & 12000                                                & iron\_axe                                                    & Plains/Forest & 0.3611                                                  & 72                                                   & Smelt and craft a smithing\_table.                                                             \\
tripwire\_hook                   & 12000                                                & iron\_axe                                                    & Plains/Forest & 0.45                                                    & 60                                                   & Smelt and craft a tripwire\_hook.                                                              \\
chain                            & 12000                                                & iron\_axe                                                    & Plains/Forest & 0.3729                                                  & 59                                                   & Smelt and craft a chain.                                                                       \\
iron\_bars                       & 12000                                                & iron\_axe                                                    & Plains/Forest & 0.3208                                                  & 53                                                   & Smelt and craft an iron\_bars.                                                                 \\
hopper                           & 12000                                                & iron\_axe                                                    & Plains/Forest & 0.3077                                                  & 65                                                   & Smelt and craft a hopper.                                                                      \\
iron\_nugget                     & 12000                                                & iron\_axe                                                    & Plains/Forest & 0.3582                                                  & 67                                                   & Smelt and craft an iron\_nugget.                                                               \\
heavy\_weighted\_pressure\_plate & 12000                                                & iron\_axe                                                    & Plains/Forest & 0.358                                                   & 81                                                   & \begin{tabular}[c]{@{}l@{}}Smelt and craft a \\ heavy\_weighted\_pressure\_plate.\end{tabular} \\
blast\_furnace                   & 12000                                                & iron\_axe                                                    & Plains/Forest & 0.5                                                     & 60                                                   & Smelt and craft a blast\_furnace.                                                              \\
shears                           & 12000                                                & iron\_axe                                                    & Plains/Forest & 0.25                                                    & 64                                                   & Smelt and craft a shears.                                                                      \\
stonecutter                      & 12000                                                & iron\_axe                                                    & Plains/Forest & 0.5                                                     & 60                                                   & Smelt and craft a stonecutter.                                                                 \\
iron\_hoe                        & 12000                                                & iron\_axe                                                    & Plains/Forest & 0.3214                                                  & 56                                                   & Smelt and craft an iron\_hoe.                                                                  \\
crossbow                         & 12000                                                & iron\_axe                                                    & Plains/Forest & 0.047                                                   & 63                                                   & Smelt and craft a crossbow.                                                                    \\ \bottomrule
\end{tabular}}
\end{table}

%% file: tables/results_gold.tex
\begin{table}[]
\centering
\captionsetup{justification=centering}
\caption{The results of our agent on various tasks in the \texttt{Gold} group.}
\label{tab:gold_results}
\resizebox{0.9\linewidth}{!}{
\renewcommand\arraystretch{1.1}
\begin{tabular}{@{}lllllll@{}}
\toprule
Task            & \begin{tabular}[c]{@{}l@{}}Max.\\ Steps\end{tabular} & \begin{tabular}[c]{@{}l@{}}Initial \\ Inventory\end{tabular} & Biome         & \begin{tabular}[c]{@{}l@{}}Success \\ Rate\end{tabular} & \begin{tabular}[c]{@{}l@{}}Eval\\ Times\end{tabular} & Language Instruction               \\ \midrule
golden\_pickaxe & 36000                                                & iron\_axe                                                    & Plains/Forest & 0.0526                                                  & 77                                                   & Smelt and craft a golden\_pickaxe. \\
golden\_shovel  & 36000                                                & iron\_axe                                                    & Plains/Forest & 0.0822                                                  & 73                                                   & Smelt and craft a golden\_shovel.  \\
golden\_sword   & 36000                                                & iron\_axe                                                    & Plains/Forest & 0.0476                                                  & 85                                                   & Smelt and craft a golden\_sword.   \\
golden\_hoe     & 36000                                                & iron\_axe                                                    & Plains/Forest & 0.058                                                   & 69                                                   & Smelt and craft a golden\_hoe.     \\
golden\_axe     & 36000                                                & iron\_axe                                                    & Plains/Forest & 0.0469                                                  & 64                                                   & Smelt and craft a golden\_axe.     \\
golden\_apple   & 36000                                                & iron\_axe                                                    & Plains/Forest & 0.02                                                    & 76                                                   & Smelt and craft a golden\_apple.   \\
clock           & 36000                                                & iron\_axe                                                    & Plains/Forest & 0.02                                                    & 77                                                   & Smelt and craft a clock.           \\
gold\_nugget    & 36000                                                & iron\_axe                                                    & Plains/Forest & 0.1444                                                  & 91                                                   & Smelt and craft a gold\_nugget.    \\
gold\_ingot     & 36000                                                & iron\_axe                                                    & Plains/Forest & 0.1449                                                  & 70                                                   & Smelt and craft a gold\_ingot.     \\ \bottomrule
\end{tabular}}
\end{table}

%% file: tables/results_diamond.tex
\begin{table}[]
\centering
\captionsetup{justification=centering}
\caption{The results of our agent on various tasks in the \texttt{Diamond} group.}
\label{tab:diamond_results}
\resizebox{0.95\linewidth}{!}{
\renewcommand\arraystretch{1.1}
\begin{tabular}{@{}lllllll@{}}
\toprule
Task             & \begin{tabular}[c]{@{}l@{}}Max.\\ Steps\end{tabular} & \begin{tabular}[c]{@{}l@{}}Initial \\ Inventory\end{tabular} & Biome         & \begin{tabular}[c]{@{}l@{}}Success \\ Rate\end{tabular} & \begin{tabular}[c]{@{}l@{}}Eval\\ Times\end{tabular} & Language Instruction                                                                           \\ \midrule
diamond\_pickaxe & 36000                                                & iron\_axe                                                    & Plains/Forest & 0.0622                                                  & 692                                                  & \begin{tabular}[c]{@{}l@{}}Dig down to mine diamond and craft diamond\_pickaxe.\end{tabular} \\
diamond\_shovel  & 36000                                                & iron\_axe                                                    & Plains/Forest & 0.1136                                                  & 88                                                   & \begin{tabular}[c]{@{}l@{}}Dig down to mine diamond and craft diamond\_shovel.\end{tabular}  \\
diamond\_sword   & 36000                                                & iron\_axe                                                    & Plains/Forest & 0.1134                                                  & 97                                                   & \begin{tabular}[c]{@{}l@{}}Dig down to mine diamond and craft diamond\_sword.\end{tabular}   \\
diamond\_hoe     & 36000                                                & iron\_axe                                                    & Plains/Forest & 0.0441                                                  & 68                                                   & \begin{tabular}[c]{@{}l@{}}Dig down to mine diamond and craft diamond\_hoe.\end{tabular}     \\
diamond\_axe     & 36000                                                & iron\_axe                                                    & Plains/Forest & 0.0986                                                  & 71                                                   & \begin{tabular}[c]{@{}l@{}}Dig down to mine diamond and craft diamond\_axe.\end{tabular}     \\
diamond          & 36000                                                & iron\_axe                                                    & Plains/Forest & 0.092                                                   & 728                                                  & \begin{tabular}[c]{@{}l@{}}Dig down to mine diamond and craft diamond.\end{tabular}          \\
jukebox          & 36000                                                & iron\_axe                                                    & Plains/Forest & 0.1053                                                  & 79                                                   & \begin{tabular}[c]{@{}l@{}}Dig down to mine diamond and craft jukebox.\end{tabular}          \\ \bottomrule
\end{tabular}}
\end{table}

%% file: tables/results_redstone.tex
\begin{table}[]
\centering
\captionsetup{justification=centering}
\caption{The results of our agent on various tasks in the \texttt{Redstone} group.}
\label{tab:redstone_results}
\resizebox{0.9\linewidth}{!}{
\renewcommand\arraystretch{1.1}
\begin{tabular}{@{}lllllll@{}}
\toprule
Task            & \begin{tabular}[c]{@{}l@{}}Max.\\ Steps\end{tabular} & \begin{tabular}[c]{@{}l@{}}Initial \\ Inventory\end{tabular} & Biome         & \begin{tabular}[c]{@{}l@{}}Success \\ Rate\end{tabular} & \begin{tabular}[c]{@{}l@{}}Eval\\ Times\end{tabular} & Language Instruction                    \\ \midrule
piston          & 36000                                                & iron\_axe                                                    & Plains/Forest & 0.1772                                                  & 79                                                   & Mine redstone and make piston.          \\
redstone\_torch & 36000                                                & iron\_axe                                                    & Plains/Forest & 0.2584                                                  & 89                                                   & Mine redstone and make redstone\_torch. \\
redstone\_block & 36000                                                & iron\_axe                                                    & Plains/Forest & 0.2469                                                  & 81                                                   & Mine redstone and make redstone\_block. \\
activator\_rail & 36000                                                & iron\_axe                                                    & Plains/Forest & 0.0159                                                  & 63                                                   & Mine redstone and make activator\_rail. \\
compass         & 36000                                                & iron\_axe                                                    & Plains/Forest & 0.0759                                                  & 79                                                   & Mine redstone and make compass.         \\
dropper         & 36000                                                & iron\_axe                                                    & Plains/Forest & 0.2278                                                  & 79                                                   & Mine redstone and make dropper.         \\
note\_block     & 36000                                                & iron\_axe                                                    & Plains/Forest & 0.2239                                                  & 67                                                   & Mine redstone and make note\_block.     \\ \bottomrule
\end{tabular}}
\end{table}

%% file: tables/results_blocks.tex
\begin{table}[]
\centering
\captionsetup{justification=centering}
\caption{The results of our agent on various tasks in the \texttt{Blocks} group.}
\label{tab:blocks_results}
\resizebox{0.9\linewidth}{!}{
\renewcommand\arraystretch{1.1}
\begin{tabular}{@{}lllllll@{}}
\toprule
Task           & \begin{tabular}[c]{@{}l@{}}Max.\\ Steps\end{tabular} & \begin{tabular}[c]{@{}l@{}}Initial \\ Inventory\end{tabular} & Biome         & \begin{tabular}[c]{@{}l@{}}Success \\ Rate\end{tabular} & \begin{tabular}[c]{@{}l@{}}Eval\\ Times\end{tabular} & Language Instruction                          \\ \midrule
diorite        & 12000                                                & iron\_axe                                                    & Plains/Forest & 0.9                                                     & 30                                                   & Dig down to mine diorite block.               \\
andesite       & 12000                                                & iron\_axe                                                    & Plains/Forest & 0.9667                                                  & 30                                                   & Dig down to mine andesite block.              \\
granite        & 12000                                                & iron\_axe                                                    & Plains/Forest & 0.8667                                                  & 30                                                   & Dig down to mine granite block.               \\
coal           & 12000                                                & iron\_axe                                                    & Plains/Forest & 0.6667                                                  & 30                                                   & Dig down to mine coal block.                  \\
lapis\_lazuli  & 12000                                                & iron\_axe                                                    & Plains/Forest & 0.8667                                                  & 30                                                   & Dig down to mine lapis\_lazuli block.         \\
iron\_ore      & 12000                                                & iron\_axe                                                    & Plains/Forest & 0.5667                                                  & 30                                                   & Dig down to mine iron\_ore block.             \\
gold\_ore      & 36000                                                & iron\_axe                                                    & Plains/Forest & 0.27                                                    & 30                                                   & Dig down to mine gold\_ore block.             \\ 
cobblestone    & 12000                                                & iron\_axe                                                    & Plains/Forest & 0.9667                                                  & 30                                                   & Dig down to mine cobblestone block.           \\
gravel         & 12000                                                & iron\_axe                                                    & Plains/Forest & 0.9667                                                  & 30                                                   & Dig down to mine gravel block.                \\
oak\_log       & 12000                                                & iron\_axe                                                    & Plains/Forest & 0.9667                                                  & 30                                                   & Chop down tree and mine oak\_log block.       \\
birch\_log     & 12000                                                & iron\_axe                                                    & Plains/Forest & 0.8718                                                  & 39                                                   & Chop down tree and mine birch\_log block.     \\
acacia\_log    & 12000                                                & iron\_axe                                                    & Plains/Forest & 0.5                                                     & 30                                                   & Chop down tree and mine acacia\_log block.    \\
jungle\_log    & 12000                                                & iron\_axe                                                    & Plains/Forest & 0.9333                                                  & 30                                                   & Chop down tree and mine jungle\_log block.    \\
dark\_oak\_log & 12000                                                & iron\_axe                                                    & Plains/Forest & 0.9                                                     & 30                                                   & Chop down tree and mine dark\_oak\_log block. \\
spruce\_log    & 12000                                                & iron\_axe                                                    & Plains/Forest & 0.9333                                                  & 30                                                   & Chop down tree and mine spruce\_log block.    \\ \bottomrule
\end{tabular}}
\end{table}

%% file: tables/results_armor.tex
\begin{table}[]
\centering
\captionsetup{justification=centering}
\caption{The results of our agent on various tasks in the \texttt{Armor} group.}
\label{tab:armor_results}
\resizebox{0.9\linewidth}{!}{
\renewcommand\arraystretch{1.1}
\begin{tabular}{@{}lllllll@{}}
\toprule
Task                & \begin{tabular}[c]{@{}l@{}}Max.\\ Steps\end{tabular} & \begin{tabular}[c]{@{}l@{}}Initial \\ Inventory\end{tabular} & Biome         & \begin{tabular}[c]{@{}l@{}}Success \\ Rate\end{tabular} & \begin{tabular}[c]{@{}l@{}}Eval\\ Times\end{tabular} & Language Instruction                    \\ \midrule
shield              & 12000                                                & iron\_axe                                                    & Plains/Forest & 0.3939                                                  & 66                                                   & Craft shield and equip it.              \\
leather\_helmet     & 12000                                                & iron\_axe                                                    & Plains/Forest & 0.0508                                                  & 59                                                   & Craft leather\_helmet and equip it.     \\
leather\_chestplate & 12000                                                & iron\_axe                                                    & Plains/Forest & 0.0312                                                  & 32                                                   & Craft leather\_chestplate and equip it. \\
leather\_leggings   & 12000                                                & iron\_axe                                                    & Plains/Forest & 0.0588                                                  & 34                                                   & Craft leather\_leggings and equip it.   \\
leather\_boots      & 12000                                                & iron\_axe                                                    & Plains/Forest & 0.087                                                   & 23                                                   & Craft leather\_boots and equip it.      \\
iron\_chestplate    & 12000                                                & iron\_axe                                                    & Plains/Forest & 0.3333                                                  & 30                                                   & Craft iron\_chestplate and equip it.    \\
iron\_boots         & 12000                                                & iron\_axe                                                    & Plains/Forest & 0.3667                                                  & 30                                                   & Craft iron\_boots and equip it.         \\ 
iron\_leggings      & 12000                                                & iron\_axe                                                    & Plains/Forest & 0.3788                                                  & 66                                                   & Craft iron\_leggings and equip it.      \\
iron\_helmet        & 12000                                                & iron\_axe                                                    & Plains/Forest & 0.303                                                   & 33                                                   & Craft iron\_helmet and equip it.        \\
diamond\_helmet     & 36000                                                & iron\_axe                                                    & Plains/Forest & 0.0429                                                  & 70                                                   & Craft diamond\_helmet and equip it.     \\
diamond\_chestplate & 36000                                                & iron\_axe                                                    & Plains/Forest & 0.0149                                                  & 68                                                   & Craft diamond\_chestplate and equip it. \\
diamond\_leggings   & 36000                                                & iron\_axe                                                    & Plains/Forest & 0.02                                                    & 73                                                   & Craft diamond\_leggings and equip it.   \\
diamond\_boots      & 36000                                                & iron\_axe                                                    & Plains/Forest & 0.0533                                                  & 75                                                   & Craft diamond\_boots and equip it.      \\
golden\_helmet      & 36000                                                & iron\_axe                                                    & Plains/Forest & 0.0533                                                  & 75                                                   & Craft golden\_helmet and equip it.      \\
golden\_chestplate  & 36000                                                & iron\_axe                                                    & Plains/Forest & 0.02                                                    & 78                                                   & Craft golden\_chestplate and equip it.  \\
golden\_leggings    & 36000                                                & iron\_axe                                                    & Plains/Forest & 0.0159                                                  & 89                                                   & Craft golden\_leggings and equip it.    \\
golden\_boots       & 36000                                                & iron\_axe                                                    & Plains/Forest & 0.0617                                                  & 81                                                   & Craft golden\_boots and equip it.       \\ \bottomrule
\end{tabular}}
\end{table}

%% file: tables/results_decoration.tex
\begin{table}[]
\centering
\captionsetup{justification=centering}
\caption{The results of our agent on various tasks in the \texttt{Decoration} group.}
\label{tab:decoration_results}
\resizebox{0.85\linewidth}{!}{
\renewcommand\arraystretch{1.1}
\begin{tabular}{@{}lllllll@{}}
\toprule
Task              & \begin{tabular}[c]{@{}l@{}}Max.\\ Steps\end{tabular} & \begin{tabular}[c]{@{}l@{}}Initial \\ Inventory\end{tabular} & Biome         & \begin{tabular}[c]{@{}l@{}}Success \\ Rate\end{tabular} & \begin{tabular}[c]{@{}l@{}}Eval\\ Times\end{tabular} & Language Instruction          \\ \midrule
yellow\_dye       & 12000                                                & iron\_axe                                                    & Flower Forest & 0.2333                                                  & 30                                                   & Obtain the yellow\_dye.       \\
red\_dye          & 12000                                                & iron\_axe                                                    & Flower Forest & 0.6364                                                  & 33                                                   & Obtain the red\_dye.          \\
light\_gray\_dye  & 12000                                                & iron\_axe                                                    & Flower Forest & 0.6667                                                  & 27                                                   & Obtain the light\_gray\_dye.  \\
pink\_dye         & 12000                                                & iron\_axe                                                    & Flower Forest & 0.6667                                                  & 39                                                   & Obtain the pink\_dye.         \\
orange\_dye       & 12000                                                & iron\_axe                                                    & Flower Forest & 0.4857                                                  & 35                                                   & Obtain the orange\_dye.       \\
white\_dye        & 12000                                                & iron\_axe                                                    & Flower Forest & 0.1471                                                  & 34                                                   & Obtain the white\_dye.        \\
white\_bed        & 12000                                                & iron\_axe                                                    & Flower Forest & 0.5                                                     & 36                                                   & Obtain the white\_bed.        \\ 
item\_frame       & 12000                                                & iron\_axe                                                    & Flower Forest & 0.2143                                                  & 28                                                   & Obtain the item\_frame.       \\
painting          & 12000                                                & iron\_axe                                                    & Flower Forest & 0.5484                                                  & 31                                                   & Obtain the painting.          \\
white\_wool       & 12000                                                & iron\_axe                                                    & Flower Forest & 0.8235                                                  & 34                                                   & Obtain the white\_wool.       \\
white\_carpet     & 12000                                                & iron\_axe                                                    & Flower Forest & 0.6857                                                  & 35                                                   & Obtain the white\_carpet.     \\
white\_banner     & 12000                                                & iron\_axe                                                    & Flower Forest & 0.0968                                                  & 31                                                   & Obtain the white\_banner.     \\
yellow\_wool      & 12000                                                & iron\_axe                                                    & Flower Forest & 0.0625                                                  & 32                                                   & Obtain the yellow\_wool.      \\
red\_wool         & 12000                                                & iron\_axe                                                    & Flower Forest & 0.6571                                                  & 35                                                   & Obtain the red\_wool.         \\
light\_gray\_wool & 12000                                                & iron\_axe                                                    & Flower Forest & 0.6098                                                  & 41                                                   & Obtain the light\_gray\_wool. \\
pink\_wool        & 12000                                                & iron\_axe                                                    & Flower Forest & 0.4                                                     & 25                                                   & Obtain the pink\_wool.        \\
orange\_wool      & 12000                                                & iron\_axe                                                    & Flower Forest & 0.5                                                     & 36                                                   & Obtain the orange\_wool.      \\ \bottomrule
\end{tabular}}
\end{table}

%% file: tables/results_food.tex
\begin{table}[]
\centering
\captionsetup{justification=centering}
\caption{The results of our agent on various tasks in the \texttt{Food} group.}
\label{tab:food_results}
\resizebox{0.9\linewidth}{!}{
\renewcommand\arraystretch{1.1}
\begin{tabular}{@{}lllllll@{}}
\toprule
Task             & \begin{tabular}[c]{@{}l@{}}Max.\\ Steps\end{tabular} & \begin{tabular}[c]{@{}l@{}}Initial \\ Inventory\end{tabular} & Biome  & \begin{tabular}[c]{@{}l@{}}Success \\ Rate\end{tabular} & \begin{tabular}[c]{@{}l@{}}Eval\\ Times\end{tabular} & Language Instruction                        \\ \midrule
apple            & 12000                                                & iron\_axe                                                    & Plains & 0.5                                                     & 30                                                   & Chop down tree to obtain apple.             \\
cooked\_chicken  & 12000                                                & iron\_axe                                                    & Plains & 0.3562                                                  & 73                                                   & Kill chicken to obtain chicken and cook it. \\
cooked\_mutton   & 12000                                                & iron\_axe                                                    & Plains & 0.4355                                                  & 62                                                   & Kill sheep to obtain mutton and cook it.    \\
cooked\_porkchop & 12000                                                & iron\_axe                                                    & Plains & 0.3968                                                  & 63                                                   & Kill pig to obtain porkchop and cook it.    \\
cooked\_beef     & 12000                                                & iron\_axe                                                    & Plains & 0.2857                                                  & 63                                                   & Kill cow to obtain beef and cook it.        \\
chicken          & 12000                                                & iron\_axe                                                    & Plains & 0.5667                                                  & 30                                                   & Kill chicken to obtain chicken.             \\
beef             & 12000                                                & iron\_axe                                                    & Plains & 0.6333                                                  & 30                                                   & Kill cow to obtain beef.                    \\ 
mutton           & 12000                                                & iron\_axe                                                    & Plains & 0.5667                                                  & 30                                                   & Kill sheep to obtain mutton.                \\
porkchop         & 12000                                                & iron\_axe                                                    & Plains & 0.4667                                                  & 30                                                   & Kill pig to obtain porkchop.                \\ \bottomrule
\end{tabular}}
\end{table}